\documentclass[journal]{IEEEtran}

\usepackage{tabularx}
\usepackage{booktabs}
\usepackage{multirow}
\usepackage{amsmath}
\usepackage{amssymb}
\usepackage[noadjust]{cite}
\usepackage{xparse} 
\usepackage{amsfonts} 
\usepackage{bbm}
\usepackage[svgnames,x11names]{xcolor}
\usepackage{mathtools}
\usepackage{graphicx}
\usepackage{subcaption}
\usepackage{calc}
\usepackage{balance}
\usepackage{url}
\usepackage{algorithm}
\usepackage{algorithmic}

\usepackage{mwe}
\usepackage{duckuments}

\makeatletter
\let\NAT@parse\undefined
\makeatother

\usepackage[colorlinks]{hyperref}
\hypersetup{
     colorlinks=true,
     linkcolor=blue,
     filecolor=blue,
     citecolor=Blue4,      
     urlcolor=cyan,
     }
\usepackage[capitalize]{cleveref}

\NewDocumentCommand\bbm{}{ \begin{bmatrix} }
\NewDocumentCommand\ebm{}{ \end{bmatrix} }
\NewDocumentCommand\Vector{m}{ \boldsymbol{\mathbf{#1}} }
\NewDocumentCommand\Matrix{m}{ \boldsymbol{\mathbf{#1}} }
\NewDocumentCommand\Mat{m}{ \boldsymbol{\mathbf{#1}} }

\NewDocumentCommand\Real{}{ \mathbb{R} }

\newcommand{\mdpm}{\mathcal{M}}
\newcommand{\mdps}{\mathcal{S}}

\newcommand{\mdpa}{\mathcal{A}}

\newcommand{\mdpp}{\mathcal{P}}
\newcommand{\mdprho}{\rho_0}

\NewDocumentCommand\Force{}{\boldsymbol{\mathcal{F}}}

\newcommand\hl[1]{%
  \bgroup
  \hskip0pt\color{blue!90!black}%
  #1%
  \egroup
}

\newif\ifhighlight

\ifhighlight
\else
    \renewcommand\hl[1]{#1}
\fi

\title{Multimodal and Force-Matched Imitation Learning with a See-Through Visuotactile Sensor}

\author{Trevor Ablett$^{1,2}$, Oliver Limoyo$^{1,2}$, Adam Sigal$^{2}$, Affan Jilani$^{2,3}$, \\
Jonathan Kelly$^{1}$, Kaleem Siddiqi$^{2,3}$, Francois Hogan$^{2}$, and Gregory Dudek$^{2,3}$%
\thanks{$^{1}$Authors are with the Space and Terrestrial Autonomous Robotics Systems (STARS) Laboratory and the Robotics Institute (RI) at the University of Toronto, Toronto, ON M5S 1A4, Canada. Email: first-name.last-name@robotics.utias.utoronto.ca}%
\thanks{$^{2}$Authors were with Samsung AI Centre, Montreal, QC M3H 5T6, Canada when this work was completed.}%
\thanks{$^{3}$Authors are with McGill University, Montreal, QC H3A 089, Canada. Email: first-name.last-name@mail.mcgill.ca}%
}

\begin{document}
\maketitle

\begin{abstract}

Contact-rich tasks continue to present many challenges for robotic manipulation.
In this work, we leverage a multimodal visuotactile sensor within the framework of imitation learning (IL) to perform contact-rich tasks that involve relative motion (e.g., slipping and sliding) between the end-effector and the manipulated object.
We introduce two algorithmic contributions, \textit{tactile force matching} and \textit{learned mode switching}, as complimentary methods for improving IL.
Tactile force matching enhances kinesthetic teaching by reading approximate forces during the demonstration and generating an adapted robot trajectory that recreates the recorded forces.
Learned mode switching uses IL to couple visual and tactile sensor modes with the learned motion policy, simplifying the transition from reaching to contacting.
We perform robotic manipulation experiments on four door-opening tasks with a variety of observation and algorithm configurations to study the utility of multimodal visuotactile sensing and our proposed improvements.
Our results show that the inclusion of force matching raises average policy success rates by 62.5\%, visuotactile mode switching by 30.3\%, and visuotactile data as a policy input by 42.5\%, emphasizing the value of see-through tactile sensing for IL, both for data collection to allow force matching, and for policy execution to enable accurate task feedback.
Project site: \url{https://papers.starslab.ca/sts-il}.  %

\end{abstract}

\begin{IEEEkeywords}
Force and Tactile Sensing, Imitation Learning, Learning from Demonstration, Deep Learning in Grasping and Manipulation
\end{IEEEkeywords}

\section{Introduction}
\label{sec:introduction}

\begin{figure}[b]
    \ifhighlight
        \captionsetup{labelfont={color=blue}}
    \fi
    \vspace{-5mm}
    \centering
    \begin{subfigure}[t]{0.4\columnwidth}
        \includegraphics[width=\textwidth]{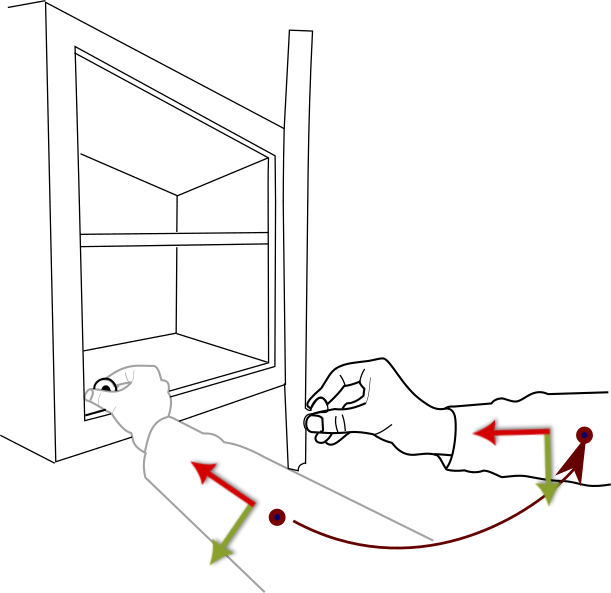}
        \caption{\hl{\textbf{Grasp and rotate.}}}
    \end{subfigure}
    \quad
    \begin{subfigure}[t]{0.4\columnwidth}
        \includegraphics[width=\textwidth]{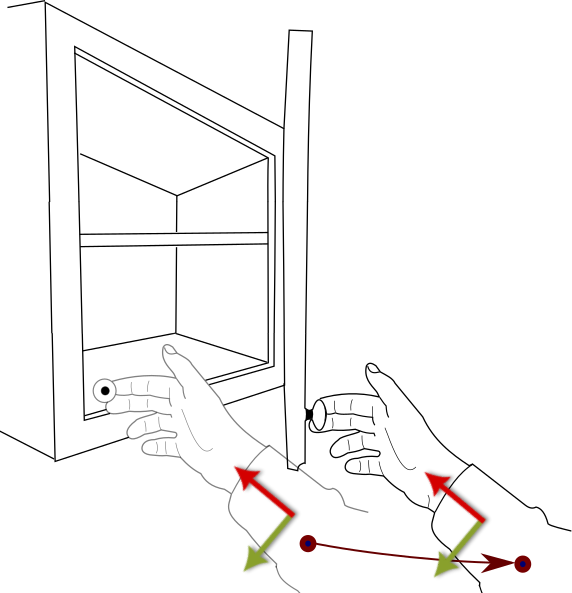}
        \caption{\hl{\textbf{Press and pull.}}}
    \end{subfigure}
    \caption{\hl{Various human approaches to opening a cabinet.
    The ``Press" approach on the right requires far less arm rotation, but also generates relative motion between the knob and the hand, motivating the use of high-resolution tactile sensing to replicate.}}
    \label{fig:grasp_press}
\end{figure}

\hl{The conventional approach to manipulating articulated objects such as doors and drawers with robots relies on  a firm, stable grasp of the handle followed by a large arm motion to complete the opening/closing task.
In contrast, humans are capable of opening and closing doors with minimal arm motions,  by relaxing their grasp on the handle and allowing for relative motion between their fingers and the handle (see \cref{fig:grasp_press}).
This paper aims to learn robot policies for door opening that are more in line with human manipulation, by leveraging high-resolution visual and tactile feedback to control the contact interactions between the robot end-effector and the handle.}

Optical tactile sensors \cite{chiRecentProgressTechnologies2018} combine a gel-based material with an internal camera to yield rich tactile information \cite{yuanGelSightHighResolutionRobot2017} and are able to provide the feedback needed for dexterous manipulation \cite{padmanabhaOmniTactMultiDirectionalHighResolution2020, maDenseTactileForce2019}.
A recently-introduced see-through-your-skin (STS) multimodal optical sensor variant combines visual sensing with tactile sensing by leveraging a transparent membrane and controllable lighting \cite{hoganSeeingYourSkin2020, hoganFingerSTSCombinedProximity2022}.
This sensor enables perception of the full interaction, from approach, through initial contact, to grasping and pulling or pushing.

\begin{figure}[t]
    \centering
    \includegraphics[width=\columnwidth]{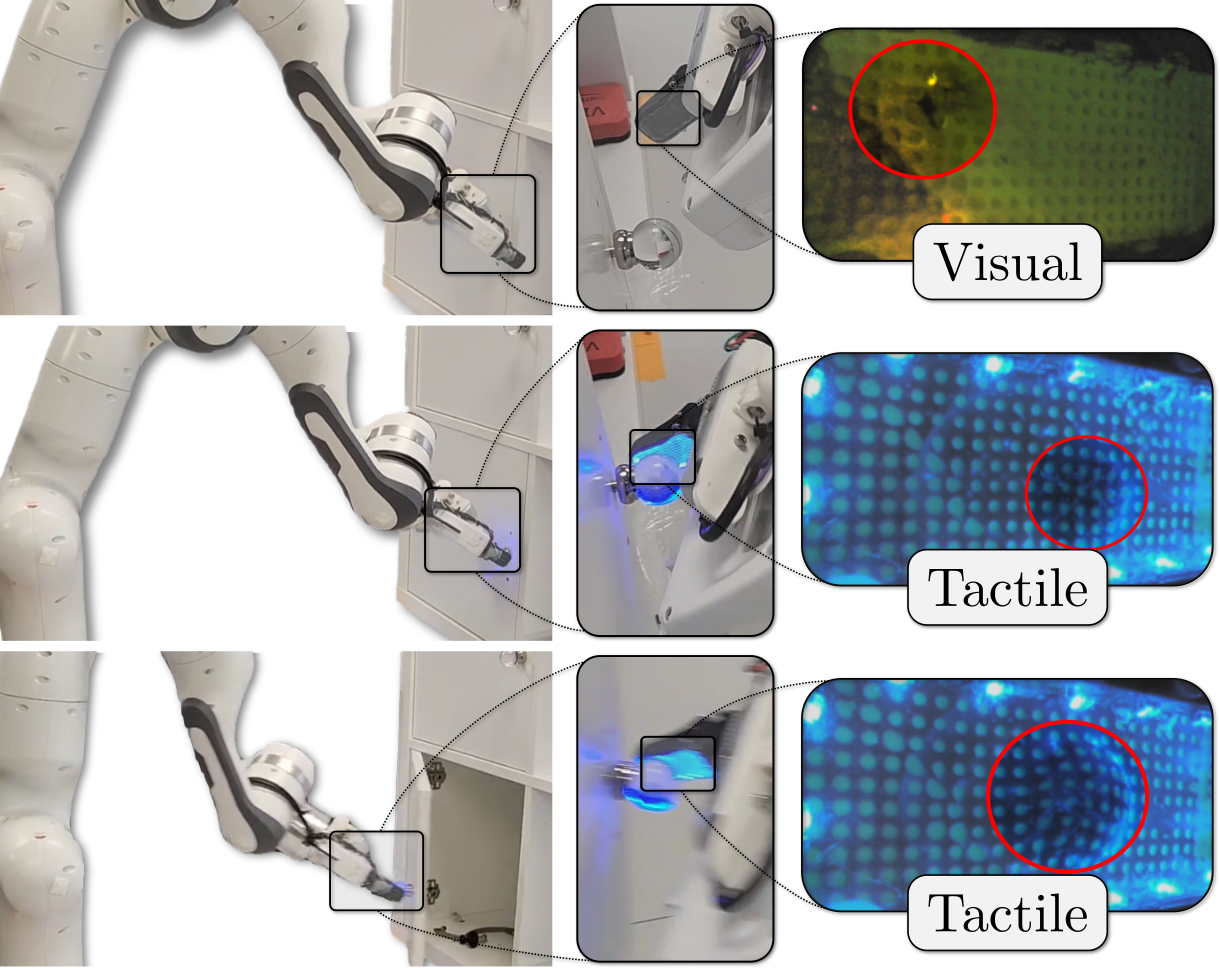}
    \caption{Our STS sensor before and during contact \hl{(right column)} with a cabinet knob \hl{(middle column)} during a door opening task \hl{(left column)}.
    \hl{In \textit{visual} mode, the camera sees through the gel membrane, allowing the knob to be found,} while \textit{tactile} mode provides contact-based feedback, via gel deformation and resultant dot displacement, upon initial contact and during opening.
    Red circles highlight the knob in the sensor view.}
    \label{fig:ex_sts}
    \vspace{-3mm}
\end{figure}

\begin{figure*}[t]
    \centering
    \includegraphics[width=\textwidth]{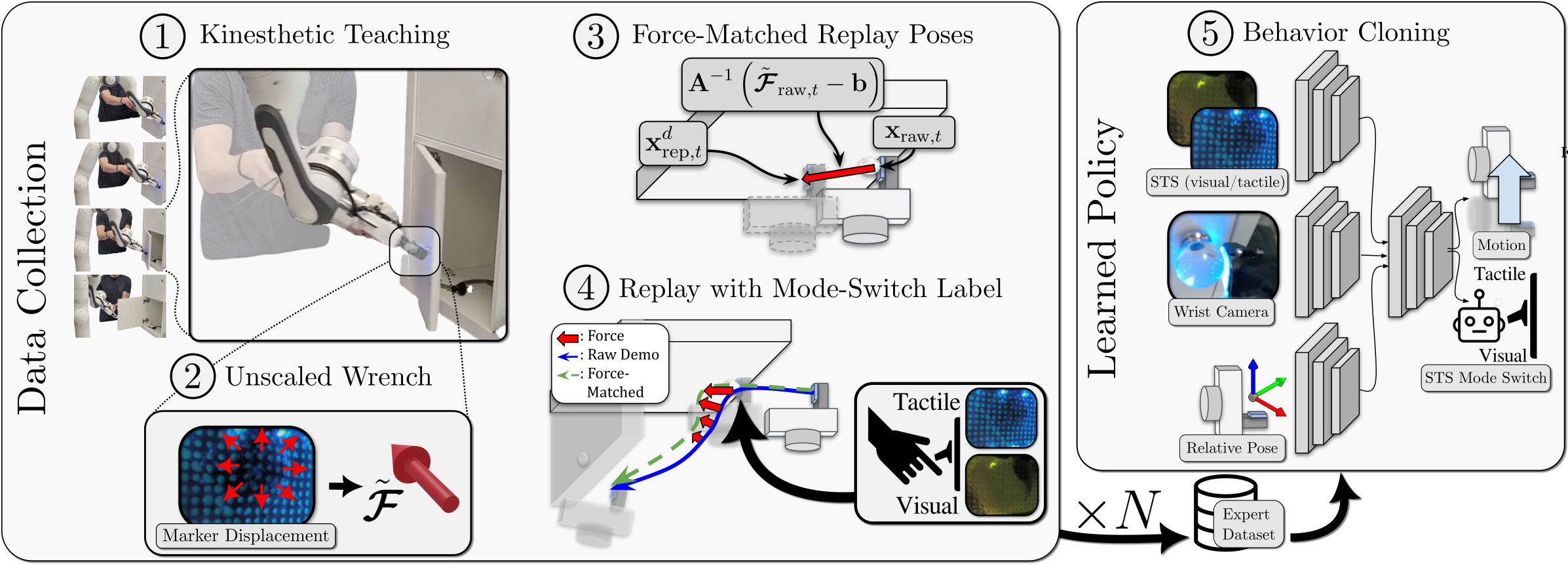}
    \caption{Visual representations of each component of our system:
    (1) Raw\hl{, human} demonstrations are \hl{generated} via kinesthetic teaching. 
    (2) During the demonstration, an STS sensor in tactile mode allows us to read a \hl{four dimensions of an unscaled wrench in $x$, $y$, $z$, and rotationally about $z$.}
    (3) \hl{For each timestep $t$ from the demonstration trajectory from (1), each raw demonstration pose $\Vector{x}_{\text{raw},t}$ uses the linear calibration parameters $\Mat{A}$ and $\Vector{b}$ (relating unscaled $\tilde{\Force}$ from (2) to control error $\Vector{e}$) and the measured wrench $\tilde{\Force}_{\text{raw},t}$ from (2) to generate a \textit{force-matched} replay pose $\Vector{x}^d_{\text{rep},t}$}.
    (4) The new, modified replay poses are used to replay the demonstration while a human provides an STS mode switch label.
    These replayed, force-matched demonstrations are stored in an expert dataset containing STS, wrist camera, and relative pose data as observations, as well as robot motion and STS mode labels as actions.
    (5) We train policies using some or all \hl{of STS, wrist camera, and relative pose data} with behavior \hl{cloning.}}
    \label{fig:system}
    \vspace{-.3cm}
\end{figure*}

\hl{In this paper, we investigate how to leverage visuotactile sensing for imitation learning (IL) on a real robotic platform for contact-rich manipulation tasks.
We focus on the tasks of opening and closing cabinet doors with challenging handle geometries (e.g., flat and spherical knobs) that are difficult to grasp with a parallel jaw gripper and that require fine motor control and tactile feedback.}
We complete these tasks with a 7-DOF robotic system that integrates a \hl{single} robotic finger outfitted with \hl{an STS} visuotactile sensor (see \cref{fig:ex_sts}), evaluating on four tasks in total.

Human-based expert demonstrations for IL can be \hl{generated} in a variety of ways, though most methods fall generally into the kinesthetic teaching (in which a person directly moves and pushes the arm to complete a task) or teleoperation (in which a person remotely controls the robot through a secondary apparatus) categories \cite{billardLearningHumans2016}.
While neither method is the definitive choice in all cases, kinesthetic teaching offers two specific, major advantages over teleoperation: 1) a degree of haptic feedback is provided to the demonstrator, since they indirectly feel contact between the end-effector and the environment (similar to tactile feedback that humans feel during tool use), and 2) no extra devices beyond the arm itself are required.
\hl{Teleoperation} requires a proxy \hl{(i.e., a separate sensor, actuator and system) to provide a substitute for true haptic feedback} \cite{ablettSeeingAllAngles2021, liImmersiveDemonstrationsAre2023}, and can be costly \hl{or} inaccurate.
Additionally, prior work has found that kinesthetic teaching is preferred over teleoperation for its ease of use and speed of providing demonstrations \cite{pervezNovelLearningDemonstration2017, fischerComparisonTypesRobot2016, akgunRobotLearningDemonstration2011}.

Unfortunately, kinesthetic teaching methods typically only measure \hl{robot} motion, without considering the \hl{robot-environment contact force (and torque)}.
To match the force profile\footnote{
\hl{We refer to forces in this section, but the method also applies to wrenches.
When necessary, we specify individual dimensions of force and torque.}
}
\hl{of the demonstration}, we require a means of measuring \hl{robot-environment forces}, in addition to a mechanism for reproducing those forces.
Our first contribution is a \textit{tactile force matching} method that uses the readings from an STS sensor to modify the poses recorded from kinesthetic teaching.
Our method generates \hl{a new trajectory that,} when used as input to a Cartesian impedance controller, recovers the recorded forces \textit{and} poses to generate a \textit{force-matched replay} (see steps 2, 3 and 4 from \hl{\cref{fig:system}).}
\hl{Assuming a linear relationship between surface deformation and force, we measure approximate force in $x$, $y$, $z$, and torque ${\tau}_z$ using tracked dot motion.}
This affords multiple advantages compared with typical approaches to force-torque sensing: (i) standard methods for measuring robot-environment force are corrupted by human-robot force (see \cref{fig:fbd}), (ii) an STS is an order of magnitude less expensive than a similarly-mounted force-torque sensor, and (iii) the STS can be additionally used, in both visual and tactile modes, to provide \hl{raw sensor data} for learning policies.

To take advantage of both visual and tactile modes of an STS sensor, a method is required to decide when to switch modes.
Previous work \cite{hoganFingerSTSCombinedProximity2022} accomplished this using \hl{known object information and a hand-crafted rule.
Our second contribution is a novel method for switching between the visual and tactile modes of an STS sensor.}
We include mode switching as a policy output, allowing the \hl{human} to set the sensor mode during the demonstration \hl{replay}, which acts as a label for the expert dataset (see step 4 of \cref{fig:system}).
We find that this approach effectively learns to switch the sensor at the point of contact, \hl{significantly improving policy performance compared with single-mode sensing.}
\hl{Our third contribution} is an extensive experimental \hl{evaluation of the} benefits of including STS visual and tactile data (depending on the active mode) as inputs to a multimodal control policy. 
We compare the use of an STS sensor (both with and without mode switching) with the use of an eye-in-hand camera.
Together, our contributions significantly improve performance on contact-rich manipulation tasks by leveraging an STS sensor.

\section{Related Work}
\label{sec:related}

\textbf{Impedance Control.} Impedance control is an approach to robotic control in which force and position are related by \hl{the dynamics of} a theoretical mass-spring-damper \hl{system \cite{hoganImpedanceControlApproach1984}.}
Impedance control can be easier to employ in robotic manipulation than \hl{standard} force control \cite{sicilianoForceControl2009} because \hl{it allows for position control} and because the contact dynamics between the robot and the environment are often difficult to model.
\hl{It is not possible to apply} desired forces in this scheme, and hybrid position/force control must be used \hl{instead \cite{raibertHybridPositionForce1981}.}

\textbf{Visuotactile Sensors.} The development of gel-based optical tactile sensors has led to a range of research \hl{on tactile} feedback \cite{yuanGelSightHighResolutionRobot2017, padmanabhaOmniTactMultiDirectionalHighResolution2020, maDenseTactileForce2019}.
\hl{Using} a semi-transparent polymer on top of these sensors allows for both visual and tactile sensing \cite{yamaguchiImplementingTactileBehaviors2017}, and is further improved through the addition of controllable lighting \cite{hoganSeeingYourSkin2020}.
These sensors can be used to \hl{map normal and shear forces to displacements by tracking printed dots embedded in the sensor membrane \cite{yuanTactileMeasurementGelSight2014, maDenseTactileForce2019, kimUVtacSwitchableUV2022}.}
\hl{In this work, we determine the approximate applied wrench based on tracked marker motion.
For example, in the $z$ dimension, we use the average surface depth change, using an algorithm that estimates the depth via dot displacement \cite{jilaniTactileRecoveryShape2024}.}

\textbf{Imitation Learning.} Imitation learning (IL) is an approach for training a control policy given a set of expert demonstrations of the desired behaviour.
IL can generally be separated into methods based on behavioural cloning \cite{bainFrameworkBehaviouralCloning1996}, in which supervised learning is carried out on the expert demonstration set, or inverse reinforcement learning \cite{abbeelApprenticeshipLearningInverse2004}, where the expert's reward function is inferred from the demonstration set.
Both behavioural cloning \cite{ablettSeeingAllAngles2021, mandlekarWhatMattersLearning2021, zhangDeepImitationLearning2018} and inverse reinforcement learning \cite{ablettLearningGuidedPlay2023, changImitationLearningObservation2023, orsiniWhatMattersAdversarial2021} have been used successfully for many robotic manipulation tasks.
Recent applications tend to avoid the use of kinesthetic teaching \cite{billardLearningHumans2016} in favour of teleoperation, despite the ability of the former to provide a degree of haptic feedback to the demonstrator.
Teleoperation requires a proxy system to provide haptic feedback \cite{ablettSeeingAllAngles2021, liImmersiveDemonstrationsAre2023} that may be inaccurate or expensive.
Although we do not compare teleoperation to kinesthetic teaching in our research, previous work has shown kinesthetic teaching to be preferred for many tasks for its ease of use and speed of demonstration \cite{pervezNovelLearningDemonstration2017, fischerComparisonTypesRobot2016, akgunRobotLearningDemonstration2011}.

Prior work that combines kinesthetic teaching with force profile reproduction \hl{does not externally measure forces \cite{leeLearningForcebasedManipulation2015, abu-dakkaForcebasedVariableImpedance2018}, or} requires one demonstration for positions and a separate demonstration for forces \cite{kormushevImitationLearningPositional2011},  which can be inconvenient and difficult to provide.
\hl{In our work, a single demonstration provides both the desired poses and forces.}

\textbf{Visuotactile Sensing and Robotic Manipulation.} 
Learning-based manipulation has benefited from the use of \hl{force-torque} sensing \cite{chebotarLearningRobotTactile2014, limoyoLearningSequentialLatent2023}, and \hl{visuotactile} sensing for both reinforcement learning \cite{hansenVisuotactileRLLearningMultimodal2022} and imitation learning \cite{huangRobotLearningDemonstration2020, liSeeHearFeel2022}.
Our learning architecture is similar to \cite{liSeeHearFeel2022}, in which \hl{tactile and visual} data are \hl{fed to a single neural network that is trained with behavioural cloning.}
In \cite{liSeeHearFeel2022}, demonstrations are \hl{generated} primarily with \hl{task-specific scripts, while we use kinesthetic teaching} without any task-based assumptions or scripted policies.
\hl{Our work is closely related to \cite{hoganFingerSTSCombinedProximity2022}, in which a multimodal tactile sensor is used to complete a bead-maze task with a robotic manipulator.
Unlike \cite{hoganFingerSTSCombinedProximity2022}, we learn a single, unified policy for motion and sensor mode-switching instead of relying on two separate, hand-crafted policies.}

\section{Methodology}
\label{sec:method}

In this section, we introduce each component of our system.
\hl{We first provide a brief background on imitation learning (\cref{sec:background}), Cartesian impedance control (\cref{sec:impedance_control_background}), and kinesthetic teaching (\cref{sec:data_collection}).}
\hl{We then present our methods for 
(i) \textit{force matching}, to match expert demonstrator wrenches (\cref{sec:method_force_matching}),
(ii) \textit{measuring unscaled forces}, where we explain how we use tactile displacement fields to provide wrench estimates (\cref{sec:method_force_measuing}),
(iii) \textit{tactile force matching}, where we implement force matching using tactile wrench estimates (\cref{sec:signal_proportional_to_force}), and
(iv) \textit{contact mode labelling}, to supervise the visuotactile modality during data collection (\cref{sec:method_mode_labelling}).}
Finally, we provide our training objective in \cref{sec:method_policy_training}.

\subsection[Markov Decision Processes and Imitation Learning]{\hl{Markov Decision Processes and Imitation Learning}}
\label{sec:background}

A Markov decision process (MDP) is a tuple $\mdpm = \langle \mdps, \mdpa, R, \mdpp, \mdprho \rangle$, where the sets $\mdps$ and $\mdpa$ are respectively the state and action space, $R : S \times A \rightarrow \Real$ is a reward function, $\mdpp$ is the state-transition environment dynamics distribution and $\mdprho$ is the initial state distribution.
\hl{A deterministic policy $\pi(s)$ generates actions $a$.}
The policy $\pi$ interacts with the environment to yield the experience $\left( s_t, a_t, r_t, s_{t + 1} \right)$ for $t = 0, \dots, T$ where $s_0 \sim \mdprho(\cdot), \hl{a_t = \pi(s_t)}, s_{t+1} \sim \mdpp(\cdot | s_t, a_t), r_t = R(s_t, a_t)$, $T$ is the finite horizon length, and $\tau_{t:T} = \{(s_t, a_t), \dots, (s_T, a_T)\}$ is the trajectory starting with $(s_t, a_t)$.

We focus on imitation learning (IL), where $R$ is unknown during training and instead we are given a finite set of task-specific expert demonstration pairs $(s,a)$, $\mathcal{D}_E = \left\{ (s,a), \dots \right\}$.
\hl{Our goal is to learn a policy that maximizes task performance on the same evaluation function used to generate the demonstrations, for exmaple, whether a door is fully opened.}

\subsection{Impedance Control}
\label{sec:impedance_control_background}

\hl{During data collection and policy execution, we control the robot arm using Cartesian impedance control. 
This strategy enables us to apply predictable forces on the environment by controlling the stiffness, damping, and desired position of the robot end effector in Cartesian space. 
By adjusting these parameters, we can regulate the interaction forces between the robot and its environment while maintaining the desired position of the end effector~\cite{hoganImpedanceControlApproach1984}.

Consider a robot arm with the motion equation
\begin{equation}
\label{eqn:uncontrolled}
    \Force = \Matrix{\Lambda}(\Vector{q}) \ddot{\Vector{x}} + \Vector{\mu}(\Vector{x},\dot{\Vector{x}}) + \Vector{\gamma}(\Vector{q}) + \Vector{\eta}(\Vector{q}, \dot{\Vector{q}}) + \Force_{\text{ext}},
\end{equation}
where $\dim \Force = \dim \Vector{x} =$ 6, $\Force$ is task-space wrench, $\Vector{q}$ is joint position, $\Vector{x}$ is task-space pose (where rotations are treated as rotation vectors), $\Matrix{\Lambda}$ is the task-space inertia matrix, $\Vector{\mu}$ is the generalized Coriolis and centrifugal force, $\Vector{\gamma}$ is the gravitation, $\Vector{\eta}$ represents further non-linear terms, and $\Force_{\text{ext}}$ are environmental contacts.
An impedance control law can be defined by
\begin{multline}
\label{eqn:controlled_with_smd}
    \Force = \Matrix{K}\Vector{e} + \Matrix{D}\dot{\Vector{e}} + \Matrix{\hat{\Lambda}}(\Vector{q}) \ddot{\Vector{x}}^d + \Vector{\hat{\mu}}(\Vector{x},\dot{\Vector{x}})\\
    + \Vector{\hat{\gamma}}(\Vector{q}) + \Vector{\hat{\eta}}(\Vector{q}, \dot{\Vector{q}}) + \Force_{\text{ext}},
\end{multline}
where $\Vector{x}^d$ is desired task-space pose, $\Vector{e} = \Vector{x}^d - \Vector{x}$ is the task-space error, $\Matrix{K}$ and $\Matrix{D}$ are the selected task-space stiffness and damping matrices, and $\Matrix{\hat{\Lambda}}, \Vector{\hat{\mu}}, \Vector{\hat{\gamma}}$ and $\Vector{\hat{\eta}}$ are the internal models of corresponding terms from \cref{eqn:uncontrolled}.
Substituting \cref{eqn:controlled_with_smd} into \cref{eqn:uncontrolled}, we arrive at the closed-loop dynamics given by
\begin{equation}
\label{eqn:closed_loop}
    \Force_{\text{ext}} = \Matrix{K}\Vector{e} + \Matrix{D}\dot{\Vector{e}} + \Matrix{\Lambda}\ddot{\Vector{e}}.
\end{equation}}

\subsection{Data Collection with Kinesthetic Teaching}
\label{sec:data_collection}
In this section, we \hl{explain} our method for \hl{generating} raw demonstrations via kinesthetic teaching using impedance control, and why this method motivates force matching.
We collect one expert dataset $\mathcal{D}_E$ for each task separately using kinesthetic teaching, \hl{where} the expert physically pushes the robot to \hl{generate} demonstrations \cite{billardLearningHumans2016} (see left side of \cref{fig:system} for an example).
To allow for demonstrations via kinesthetic teaching, we set $\Mat{K}$ and $\Mat{D}$ from \cref{eqn:closed_loop} very close to zero, ensuring the robot has full compliance with the environment.
We then record end-effector poses $\Vector{x}_{E}$ at a fixed rate as the robot is moved by the human, and use these poses, or, equivalently, the changes between poses, as expert actions \hl{\cite{billardLearningHumans2016}}. %
\hl{This recording process suffers from two limitations, however}: 
(i) the \hl{recorded states and actions} may not accurately reflect $\mdps$ and $\mdpa$, and (ii) the \hl{recorded trajectory} is unable to replicate the \hl{reference forces} generated by the \hl{human during robot-environment contact.}
An example the first case above for $\mdps$ would be the presence of the human demonstrator, or even the shadow of the human demonstrator, in the frame of a camera being used to generate $\mdps$.
For $\mdpa$, it is difficult to guarantee that the controller can \hl{accurately} reproduce the motion generated under full compliance while the human is pushing the arm.

This issue can be resolved with \textit{replays}, in which the demonstrator \hl{generates} a single demonstration trajectory $\tau_{x, \text{raw}} = \{ \Vector{x}_{\text{raw}, 0}, \dots, \Vector{x}_{\text{raw}, T} \}$, resets the environment to the same $s_0$, and then uses a sufficiently accurate controller to reproduce each $\Vector{x}_{\text{raw}}$ \cite{dasariRB2RoboticManipulation2021}.
For trajectories in free space or where minimal force is exerted on the environment, this can be enough to learn effective policies \cite{dasariRB2RoboticManipulation2021, figueroaEasykinestheticrecording2023}.
\hl{The resolution of} the second limitation requires additional sensory input, as we discuss in the next section.

\subsection{Force Matching}
\label{sec:method_force_matching}
In this section, we explain our method for generating \textit{force-matched} replays of our raw kinesthetic teaching trajectories under the assumption \hl{that the true external end-effector contact wrench can be measured.}

\hl{Assuming static equilibrium with $\dot{\Vector{e}} = \ddot{\Vector{e}} = 0$, \cref{eqn:closed_loop} simplifies to
\begin{equation}
\label{eqn:control_law_general}
    \Force = \Mat{K} \Vector{e}^x,
\end{equation}
where the position and external force are related as a spring and where we have dropped $(\cdot)_\text{ext}$ from external wrench $\Force$ for notational convenience.
In our case, $\dim \Force = \dim \Vector{e}^x = $ 6, since we control the robot in six degrees of freedom (three translational, three rotational).
We treat the rotational components of $\Vector{e}^x$ as a rotation vector.}  %
On our robot, we control joint torques\footnote{
Note that our symbol for joint torques $\boldsymbol{\tau}$ \hl{(in bold)} does not refer to a vector form of a trajectory $\tau$. 
We choose these symbols to be consistent with existing literature on both force control and imitation learning.}
$\boldsymbol{\tau}$, \hl{such that
\begin{equation}
\label{eqn:joint_torques_control}
    \boldsymbol{\tau} = \mathbf{J}^{\top} \Force,
\end{equation}
where $\mathbf{J}$ is the the current manipulator Jacobian.}
\hl{Our per-timestep discrete control setpoints are
\begin{align}
    \label{eqn:control_law_static}
    \Force_{t} & = \Mat{K} \left( \Vector{x}^d_{t} - \Vector{x}_{t} \right),\\
    \label{eqn:control_law_joint_torques}
    \boldsymbol{\tau}^d_{t} & = \mathbf{J}^{\top} \Force_{t}.
\end{align}}
Substituting \cref{eqn:control_law_static} into \cref{eqn:control_law_joint_torques} yields
\begin{equation}
\label{eqn:control_law_joint_torques_no_force}
    \boldsymbol{\tau}^d_{t} = \mathbf{J}^{\top} \Mat{K} \left( \Vector{x}^d_{t} - \Vector{x}_{t} \right),
\end{equation}
illustrating that measuring \hl{external} end-effector contact \hl{wrenches} is not \hl{necessary for} the indirect control scheme employed by a standard impedance \hl{controller \cite{villaniForceControl2016}.}

Consider a raw demonstrator trajectory of recorded end-effector poses $\tau_{x, \text{raw}} = \{ \Vector{x}_{\text{raw}, 0}, \dots, \Vector{x}_{\text{raw}, T} \}$ and wrenches $\tau_{f,\text{raw}} = \{ \Force_{\text{raw}, 0}, \dots, \Force_{\text{raw}, T} \}$ as a set of poses and wrenches that we would like our controller to achieve.
We invert the spring relationship in \cref{eqn:control_law_static} to show how we can instead solve for desired (replay) positions $\Vector{x}^d_{\text{rep}, t}$ that would generate particular wrenches $\Force_{\text{raw}, t}$ as
\begin{equation}
\label{eqn:new_desired_poses}
    \Vector{x}^d_{\text{rep}, t} = \Mat{K}^{-1} \Force_{\text{raw}, t} + \Vector{x}_{\text{raw}, t},
\end{equation}
as illustrated in the replay pose generation step of \cref{fig:system}.
\hl{Considering} the modified \textit{replay} poses from \cref{eqn:new_desired_poses} and substituting these into \cref{eqn:control_law_joint_torques_no_force} as our new desired poses, we obtain
\begin{equation}
\label{eqn:control_law_joint_torques_extra_term}
\begin{split}
    \boldsymbol{\tau}^d_{t} &= \mathbf{J}^{\top} \Mat{K} \left( \Mat{K}^{-1} \Force_{\text{raw}, t} + \Vector{x}_{\text{raw}, t} - \Vector{x}_{t} \right) \\
    &= \mathbf{J}^{\top} \Force_{\text{raw}, t} + \mathbf{J}^{\top} \Mat{K} \left( \Vector{x}_{\text{raw}, t} - \Vector{x}_{t} \right).
\end{split}
\end{equation}
\cref{eqn:control_law_joint_torques_extra_term} shows that we have modified the controller with an open-loop term \hl{to directly reproduce $\Force_{\text{raw}, t}$, assuming static conditions, while maintaining the same original stiffness/impedance control term to reproduce $\Vector{x}_{\text{raw}, t}$.}
In cases where $\Force_{\text{raw}, t} = \boldsymbol{0}$, our approach acts as a simple position-based controller.

We consider the stiffness $\Mat{K}$ to be \hl{fixed} (typically a diagonal matrix with one value for all translational components, and another for all rotational ones), selected to optimally trade off control accuracy and environmental compliance.
Using \cref{eqn:new_desired_poses}, we can generate a trajectory of replay pose setpoints $\tau^d_{x, \text{rep}} = \{ \Vector{x}^d_{\text{rep}, 0}, \dots, \Vector{x}^d_{\text{rep}, T} \}$, a new set of poses that, under static conditions, would reproduce the both the positions $\tau_{x,\text{raw}}$ and the wrenches $\tau_{f,\text{raw}}$ from the raw kinesthetic teaching \hl{trajectory.}

\hl{\subsection{Measuring Unscaled Forces with A Tactile Sensor}
\label{sec:method_force_measuing}

\begin{figure}[t]
    \ifhighlight
        \captionsetup{labelfont={color=blue}}
    \fi
    \centering
    \begin{subfigure}[t]{0.47\columnwidth}
        \includegraphics[width=\textwidth]{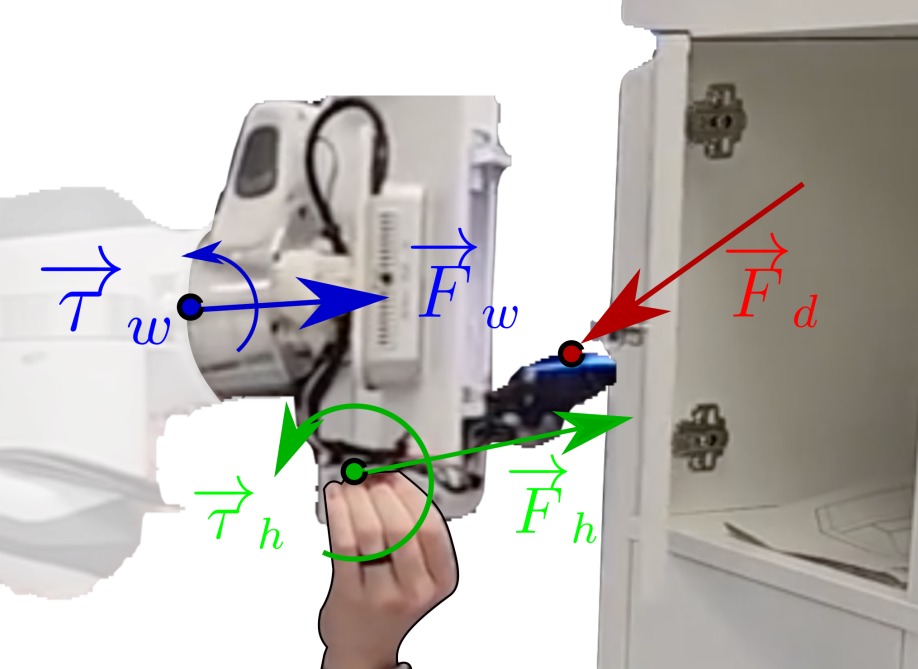}
        \caption{\hl{\textbf{Kinesthetic Teaching.}}}
    \end{subfigure}
    \hfill
    \begin{subfigure}[t]{0.47\columnwidth}
        \includegraphics[width=\textwidth]{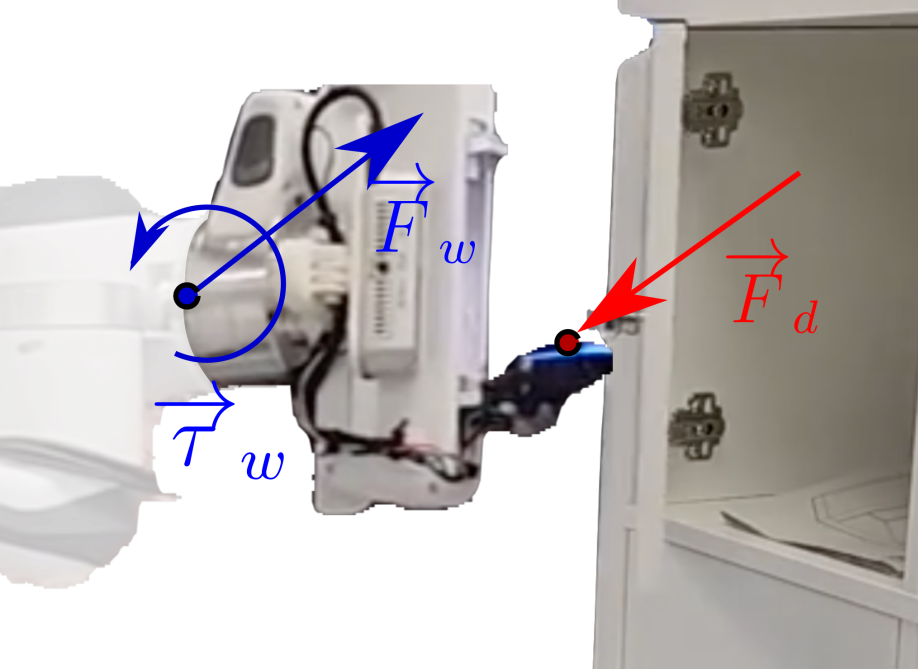}
        \caption{\hl{\textbf{Autonomous Execution.}}}
    \end{subfigure}
    \caption{\hl{With a human hand generating $\vec{F}_h$ and $\vec{\tau}_h$, wrenches measured at the wrist ($\vec{F}_w, \vec{\tau}_w$) via typical force-torque sensing modalities cannot isolate $\vec{F}_d$, as required for the force matching procedure outlined in \cref{sec:method_force_matching}. This notation applies only to this figure.}}
    \label{fig:fbd}
    \vspace{-.3cm}
\end{figure}

Our force matching method requires access to robot-environment contact wrenches.
Common approaches to measuring external end-effector wrenches include the use of a wrist-mounted force torque sensor, or kinematics and dynamics modelling combined with joint torque sensors.
\cref{fig:fbd} shows estimated wrist wrenches on a robotic end effector, and illustrates that robot-environment contact wrenches cannot be decoupled from human-generated wrenches, making wrist-measured wrench values inadequate for our purposes.
Our method requires measuring wrenches at the point of contact, making our visuotactile sensor a natural choice.
Below, we provide further detail on how we measure approximate, unscaled wrench signals $\tilde{\Force}$ with a visuotactile sensor.
In the following section (\cref{sec:signal_proportional_to_force}), we describe how to calibrate these wrench signals for use in force matching.

Prior work has shown that the relationship between surface deformation and normal force is linear for membrane-based optical tactile sensors in their elastic region \cite{yuanTactileMeasurementGelSight2014}, and as such, our approaches to approximating wrenches are based on measuring surface deformation via sensor surface marker tracking.
We track the locations of these markers in the RGB camera using OpenCV's adaptive threshold \cite{opencv_library} as well as common filtering strategies, including low-pass filtering and a scheme for rejecting outlier marker displacements based on nearest neighbour distributions.
Values for $\tilde{\mathcal{F}}_x, \tilde{\mathcal{F}}_y, \tilde{\mathcal{F}}_z$ and $\tilde{\tau}_z$ are then measured via separate methods, described below (we do not attempt to measure $\tilde{\tau}_x$ or $\tilde{\tau}_y$).}  %

For \hl{$\tilde{\mathcal{F}}_z$} (i.e., perpendicular to the surface of the finger), we use a method for estimating membrane depth, where depth is inferred from marker movement on the surface of the membrane.
\hl{The method, introduced in \cite{jilaniTactileRecoveryShape2024}, uses a} perspective camera model \hl{to recover} a relationship between the separation of markers (locally), and the amount of displacement \hl{of the surface} towards the camera.
\hl{Specifically, the Voronoi diagram between the markers and its corresponding medial axis are used to compute the changes in nearest-neighbour marker distance, giving a robust estimate of each marker's displacement towards the camera \cite{jilaniTactileRecoveryShape2024}.}
We use the average of each of these marker depths \hl{(displacements)} for the normal force $\Tilde{\mathcal{F}}_z$.
\cref{fig:calibration_normal_example} shows examples of both dot displacement and corresponding estimated depth at all points as the knob is pushed against the sensor, as well as our corresponding estimates for $\Tilde{\mathcal{F}}_z$.

For \hl{approximate} shear forces \hl{($\tilde{\mathcal{F}}_x$ and $\tilde{\mathcal{F}}_y)$}, we use the average of the tracked dot movement in both the horizontal and vertical directions parallel to the sensor plane, 
For torque \hl{(only $\tilde{\tau}_z$)}, we use the average of the tracked dot movement that is perpendicular to an estimate of a centre point of maximal normal \hl{force.}
\hl{Finally, while it would be possible to calibrate the transform between our robot and sensor frames} using standard eye-in-hand manipulator calibration \cite{tsaiNewTechniqueFully1989} or a form of touch-based eye-in-hand calibration \cite{limoyoSelfCalibrationMobileManipulator2018}, we find that assuming a fixed transformation is adequate.

\subsection[Tactile Force Matching via Calibration]{\hl{Tactile Force Matching via Calibration}}
\label{sec:signal_proportional_to_force}

\begin{figure}[b]
    \vspace{-.3cm}
    \centering
    \includegraphics[width=\columnwidth]{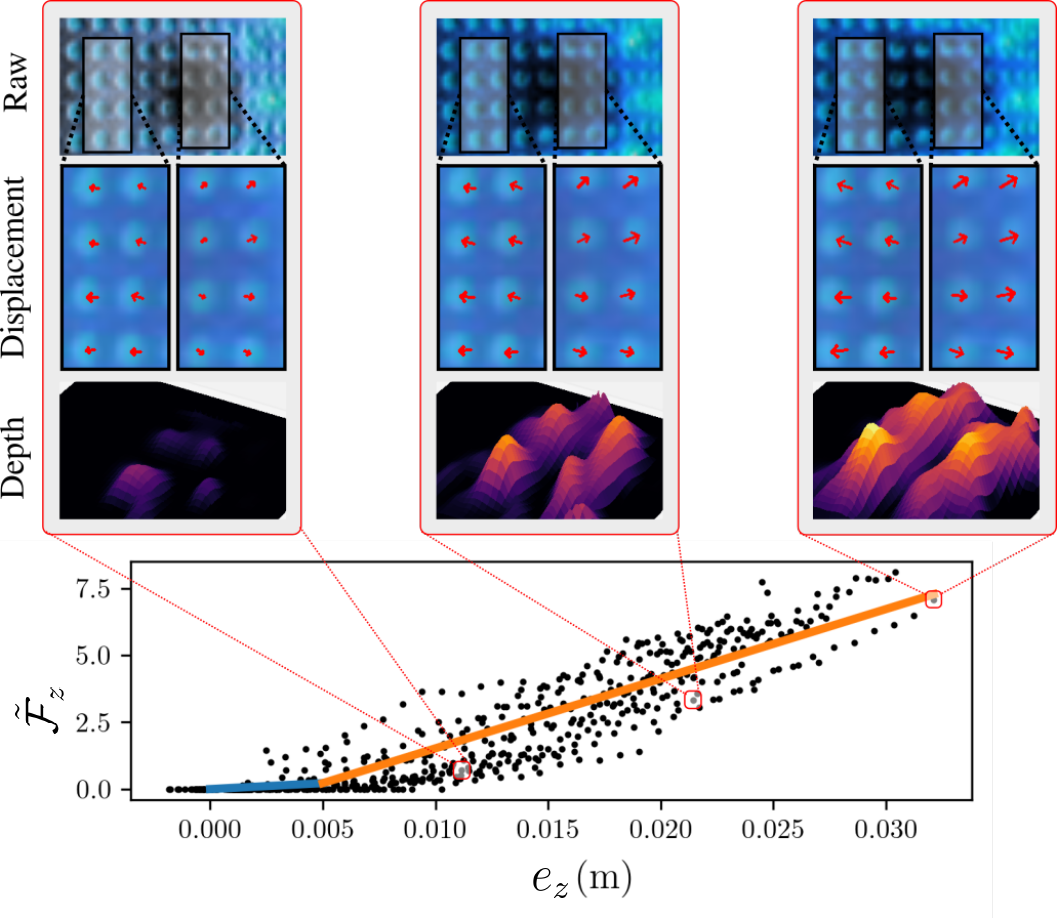}
    \caption{Example raw images \hl{along with corresponding marker displacements, inferred depths \cite{jilaniTactileRecoveryShape2024}, and $e_z$ and $\Tilde{\mathcal{F}}_z$ values}, along with the piecewise linear relationship between $e_z$ and $\Tilde{\mathcal{F}}_z$.
    See supplementary materials for corresponding video.}
    \label{fig:calibration_normal_example}
\end{figure}

\hl{As a reminder, our goal is to generate replay trajectories $\tau^d_{x, \text{rep}}$ with \cref{eqn:new_desired_poses} using a means of sensing robot-environment wrenches $\tau_{f,\text{raw}}$.
Under the assumption that the deformation signals described in \cref{sec:method_force_measuing} are linear with respect to applied wrench, and the assumption that our impedance controller adequately models the spring relationship in \cref{eqn:control_law_general}, we can directly find the relationship between our unscaled wrench $\tilde{F}$ and control position error $\Vector{e}^x = \Vector{x}^d_{t} - \Vector{x}_{t}$ by solving the linear least squares problem
\begin{equation}
\label{eqn:force_linear_relationship}
    \Tilde{\Force} = \Matrix{A} \Vector{e} + \Vector{b},
\end{equation}
where we drop the superscript $x$ from $\Vector{e}^x$ for brevity, and $\Matrix{A}$ and $\Vector{b}$ are a matrix and vector, respectively, defining the linear relationship between $\Tilde{\Force}$ and $\Vector{e}$.} %

\begin{figure*}[ht]
    \centering
    \includegraphics[width=.9\textwidth]{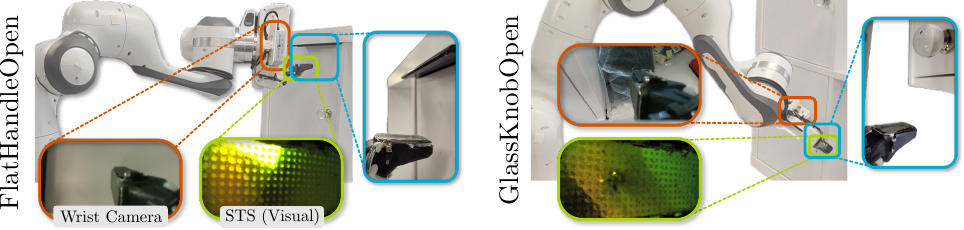}
    \caption{Our robot and sensor setup for \texttt{FlatHandleOpen} and \texttt{GlassKnobOpen}, showing example sensor data and a zoomed in view of STS and the handle/knob.
    \hl{Note that the glass knob on the top door is not used for experiments.}}
    \label{fig:env_robot_sensor_labels}
    \vspace{-.3cm}
\end{figure*}

\hl{We generate values for $\tilde{\Force}$ and $\Vector{e}$ through a short calibration procedure, where we incrementally push the sensor against the static environment numerous times, providing small initial perturbations to each trajectory to increase robustness.
The calibration procedure is not object-specific and is only performed once before completing all of our experiments.
Furthermore, we modify \cref{eqn:force_linear_relationship} to be piecewise linear, as we find that this fits our data more adequately.
See the bottom of \cref{fig:calibration_normal_example} for an example of collected $\tilde{\mathcal{F}}_z$ and $e_z$ values, as well as the corresponding piecewise linear model. 
For further description of specific design choices related to this procedure in practice, see \cref{sec:env_task_params}.}

\hl{Returning to our original goal}, we can rearrange \cref{eqn:force_linear_relationship} to resemble \cref{eqn:new_desired_poses}, yielding the individual poses in $\tau^d_{x, \text{rep}}$ as
\begin{equation}
\label{eqn:new_desired_poses_actual}
    \Vector{x}^d_{\text{rep}, t} = \Matrix{A}^{-1} \left( \Tilde{\Force}_{\text{raw}, t} - \Vector{b} \right) + \Vector{x}_{\text{raw}, t}.
\end{equation}
An advantage of this approach is that it is valid for any type of sensor \hl{that follows Hooke's law} (e.g. visuotactile, pressure, or strain-gauge sensors).
\hl{Furthermore}, any repeatable control errors (where true \hl{joint} torques do not match desired \hl{joint} torques specified in \cref{eqn:control_law_joint_torques}) or mild physical deformation of the sensor housing are accommodated, whereas simply using a \hl{finger-mounted} force-torque sensor with \cref{eqn:new_desired_poses} would not handle these issues.
We identify \hl{three potential} sources of error in this procedure: (i) measurement errors, where the same $\Vector{e}^x$ value will result in different values for $\Tilde{\Force}$, and (ii) \hl{hysteresis errors, which we do not attempt to mitigate,
and (iii) aliasing errors, where different true six-dimensional $\Force$ values project to the same four-dimensional $\tilde{\Force}$.}

\subsection{STS Sensor Mode Labelling}
\label{sec:method_mode_labelling}

\begin{figure}[b]
    \vspace{-.3cm}
    \centering
    \includegraphics[width=.9\columnwidth]{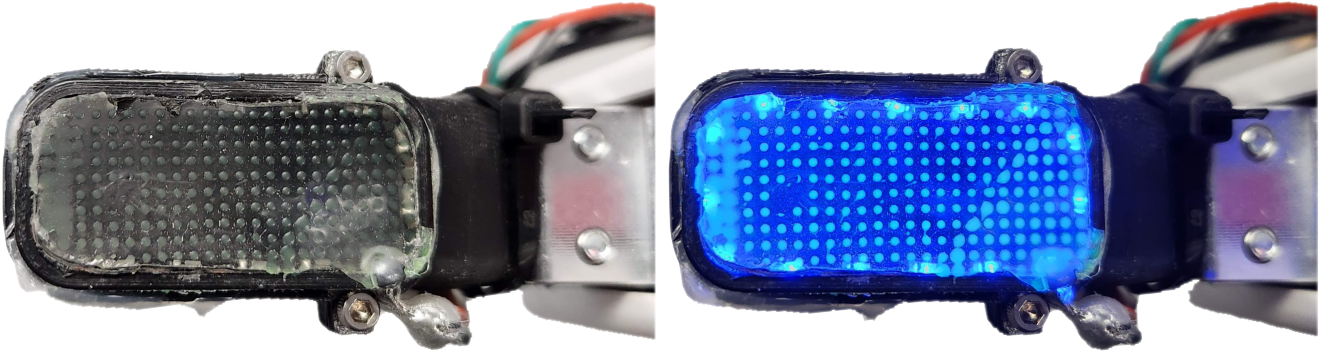}
    \caption{A front view of our finger-STS sensor in visual mode (left) and tactile mode (right), mounted on a Franka Emika Panda gripper.}
    \label{fig:sts_on_off}
\end{figure}

\hl{Unlike a regular visuotactile sensor \cite{yuanTactileMeasurementGelSight2014, maDenseTactileForce2019}, an STS sensor has a semi-transparent membrane and a ring of controllable LED lights, as shown in \cref{fig:sts_on_off}. %
The STS sensor can operate in two modes: \textit{visual mode}, where the LEDs are off, allowing for pre-contact scene observation by the camera, and \textit{tactile mode}, where the LEDs are on, enabling contact feedback.}

An STS sensor requires a method to control the switching between visual and tactile \hl{modes.}
We treat the mode switching \hl{signal} as an output from our control policy, and let the human demonstrator switch the mode of the sensor as part of their demonstration.
\hl{During the replay phase}, the controller autonomously generates $\tau_{x, \text{rep}}$, and the demonstrator observes and presses a button to change the sensor mode.
This mode change is also used as an action label for training policy outputs.
Our approach ensures that the visuotactile images that are added to the expert dataset $\mathcal{D}_E$ contain both visual and tactile data, since the initial demonstration of $\tau_{x, \text{raw}}$ had the sensor mode exclusively set to to tactile mode \hl{to read $\tilde{\Force}$.} %

\hl{An advantage of this data collection method} is that the demonstrator can choose on a task-by-task basis whether the sensor mode switch should occur before contact, at the point of contact, or after contact.
Tasks that require \hl{scene tracking} until the point of contact may benefit from having the mode switch occur post-contact, while tasks which require a delicate and less forceful touch or grasp may benefit from the \hl{opposite.}

\subsection{Policy Training}
\label{sec:method_policy_training}
\hl{The previous sections detailed how we go from the human demonstration trajectories $\tau_{x, \text{raw}}$ to force-matched replay trajectories $\tau_{x, \text{rep}}$.
The motion commands used to generate these replay trajectories (the \textit{desired} replay trajectories, $\tau^d_{x, \text{rep}}$) comprise the motion components of our actions $a$ used for training policies.}
Our policies are trained with a standard mean-squared-error behavioural cloning loss, 
\begin{equation}
    \mathcal{L}(\pi) = \sum_{(s,a) \in \mathcal{D}_E} \left(\pi(s) - a \right)^2,
\end{equation}
where $s \in \mdps$ and $a \in \mdpa$. 
$\mdps$ can include any or all of raw STS images, wrist camera images, and robot pose information, while $\mdpa$ includes motion control and sensor mode commands (see \cref{sec:env_task_params} for more details).

\section{Experiments}
\label{sec:experiments}

\begin{figure*}[ht]
    \centering
    \includegraphics[width=.95\textwidth]{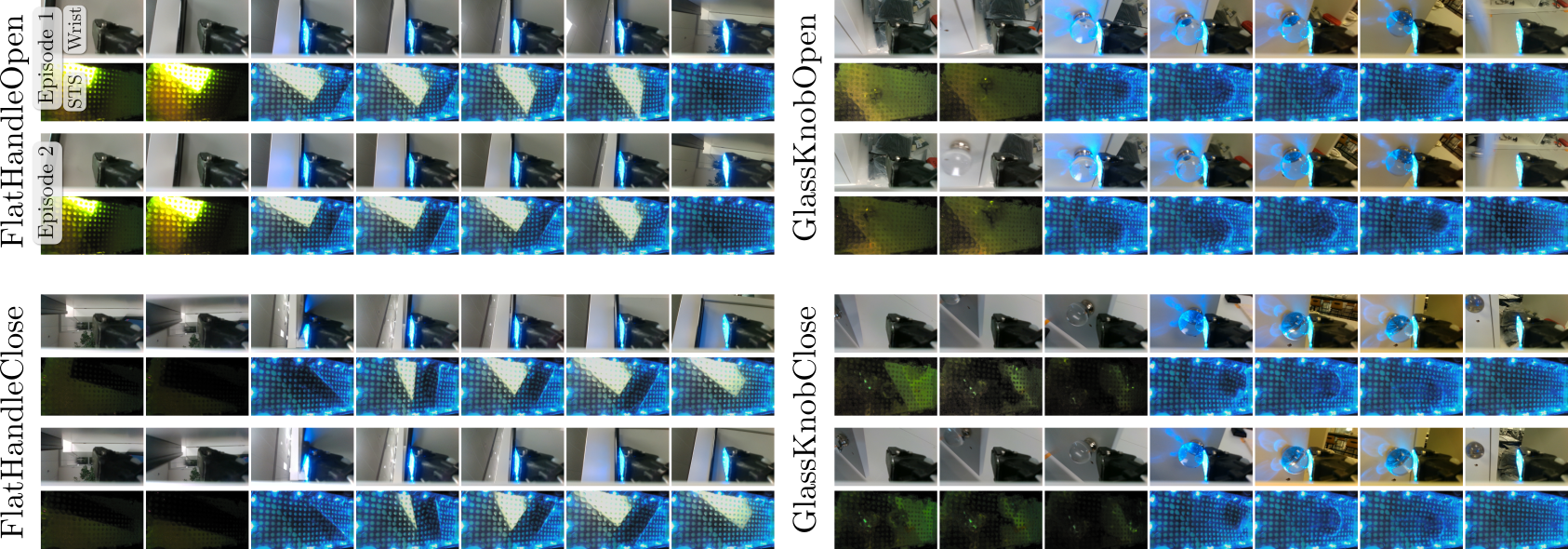}
    \caption{Wrist camera and STS data for two replayed demonstration trajectories per task (see \cref{fig:env_robot_example_trajectories} for corresponding examples showing the full scene).
    Notable features are differences in initial observations due to randomized initial poses, the change in appearance before and after the STS mode switches, the informative nature of the STS data compared with the wrist camera, and the amount of slip between the handle/knob and the sensor throughout each trajectory.}
    \label{fig:env_example_sensor_data}
    \vspace{-.3cm}
\end{figure*}

Our experimental questions are as follows.
\begin{enumerate}
    \item What are the benefits of force matching and policy mode switching for imitation learning via kinesthetic teaching?
    \item Is a see-through visuotactile sensor (STS) a required policy input for our door manipulation tasks, or is a wrist-mounted eye-in-hand camera sufficient?
    \item Can an STS sensor alone (i.e., without another external sensor) provide sufficient visual and tactile information to \hl{complete our door manipulation tasks successfully, and if so, is there a benefit to including mode switching?}
\end{enumerate}
In \cref{sec:env_task_params}, we describe our experimental environment and task parameters.
Next, we report the performance results of our system as a whole in \cref{sec:system_perf}, the benefits of using force matching in \cref{sec:force_matching_perf}, and the benefits of our policy mode-switching output in \cref{sec:policy_mode_switching_perf}.
We follow with \hl{details from our observational space study (i.e., of whether success is possible with the eye-in-hand camera alone)} in \cref{sec:exp_obs_study}.
\hl{We then give results for training and testing policies with STS data exclusively in \cref{exp:sts_only_study}.
Finally, we examine expert data scaling in \cref{sec:exp_data_scaling}.}

\subsection{Environment and Task Parameters}
\label{sec:env_task_params}

\begin{figure}[!b]
    \vspace{-.3cm}
    \centering
    \includegraphics[width=\columnwidth]{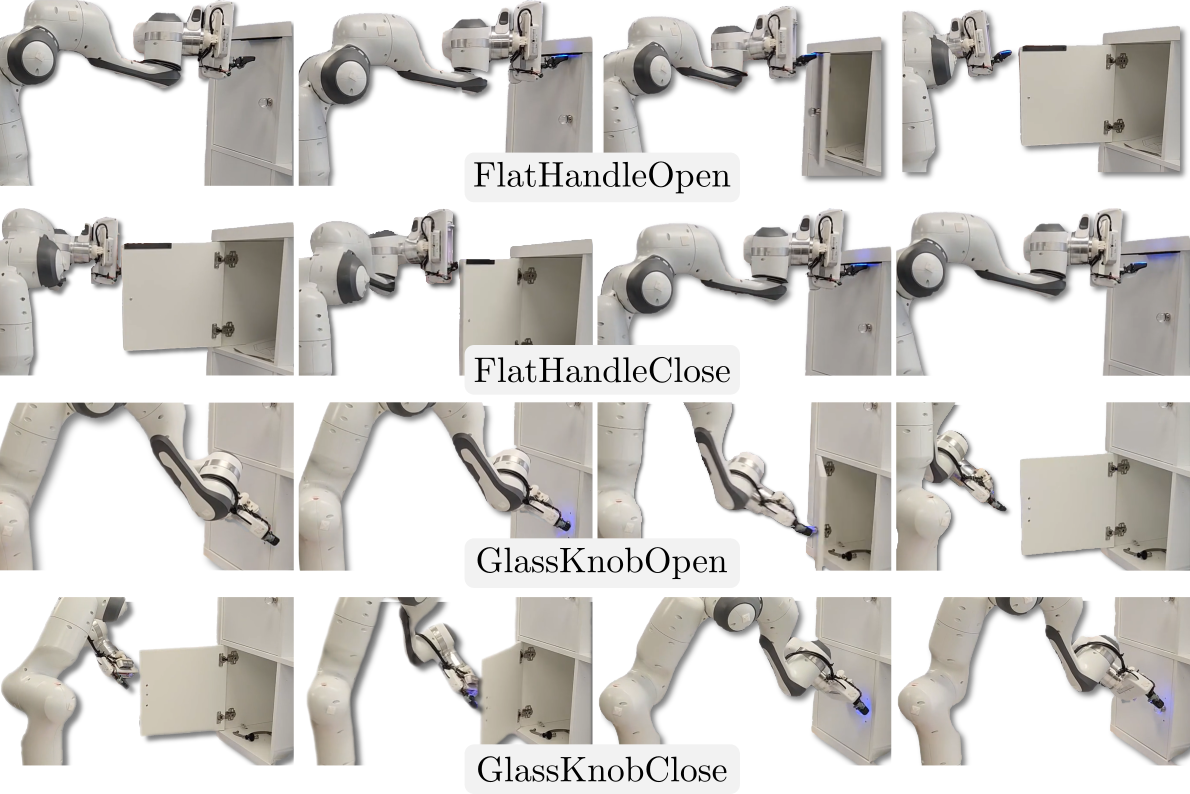}
    \caption{Example trajectories \hl{for each of our four tasks, showing the motion of the robot from approach through contact.}}
    \label{fig:env_robot_example_trajectories}
\end{figure}

We study cabinet door opening and closing, using one door with a flat metal handle and one with a spherical glass knob (see \cref{fig:env_robot_sensor_labels}), giving us four total experimental tasks, hereafter referred to as \texttt{FlatHandleOpen}, \texttt{FlatHandleClose}, \texttt{GlassKnobOpen}, and \texttt{GlassKnobClose}.
All tasks include an initial reaching/approach component (see \cref{fig:env_robot_sensor_labels}, \cref{fig:env_robot_example_trajectories}, and \cref{fig:env_example_sensor_data}), where the initial pose of the robot relative to the the knob or handle is \hl{randomized.}
See \cref{fig:env_example_sensor_data} for visual examples of the continuous slipping between the finger and the knobs/handles throughout demonstrations.

Door opening tasks are considered successful if the door fully opens within a given time limit.
Door closing tasks are considered successful if the door closes without ``slamming": \hl{if the finger fully slips off of the knob or handle before the door is closed, the spring hinges cause it to loudly slam shut.}
In either case, failure occurs because the finger loses contact with the handle or knob before the motion is \hl{complete}. %

Our robotic platform is a Franka Emika Panda, and we use the default controller that comes with Polymetis \cite{Polymetis2021} as our Cartesian impedance controller.
For all tasks, at the beginning of each episode, the initial pose of the end-effector frame is randomized to be within a 3 cm $\times$ 3 cm $\times$ 3 cm cube in free space, with the rotation about the global $z$-axis (upwards facing) randomized between $-0.15$ and $+0.15$ radians.
Our \hl{training data} can include wrist camera 212 $\times$ 120 pixel RGB images, raw 212 $\times$ 120 STS images (in either visual or tactile mode), and the current and previous relative end-effector poses (position, quaternion).
In this work, by \textit{relative poses}, we mean that, for each episode, the initial pose is set to $\Vector\{0,0,0,0,0,0,1\}$, although the \textit{global pose} is randomized for every episode, as previously described.
\hl{Finally, our action space consists of 6-DOF relative position changes in the frame of the end-effector.}
These choices are meant to simulate the situation in which an approximate reach is performed with an existing policy and global pose information between episodes \hl{is inconsistent, as is often the case in mobile manipulation \cite{ablettSeeingAllAngles2021}.}

\hl{Our sensor is based on the one described in \cite{hoganFingerSTSCombinedProximity2022,hoganSeeingYourSkin2020}, but in a smaller form factor.}
The finger housing is 3D-printed and mounted on the Franka Emika Panda gripper.
Due to the fragility of the top layer of the sensor, especially when subject to large shear forces, we use the sensor with this top layer, as well as the semi-reflective paint, fully \hl{removed.}
The LED values and camera parameters for the sensor are set using a \hl{simple tool to optimize scene visibility in visual mode, and marker visibility in tactile mode}.

\begin{figure*}[ht]
    \centering
    \includegraphics[width=\textwidth]{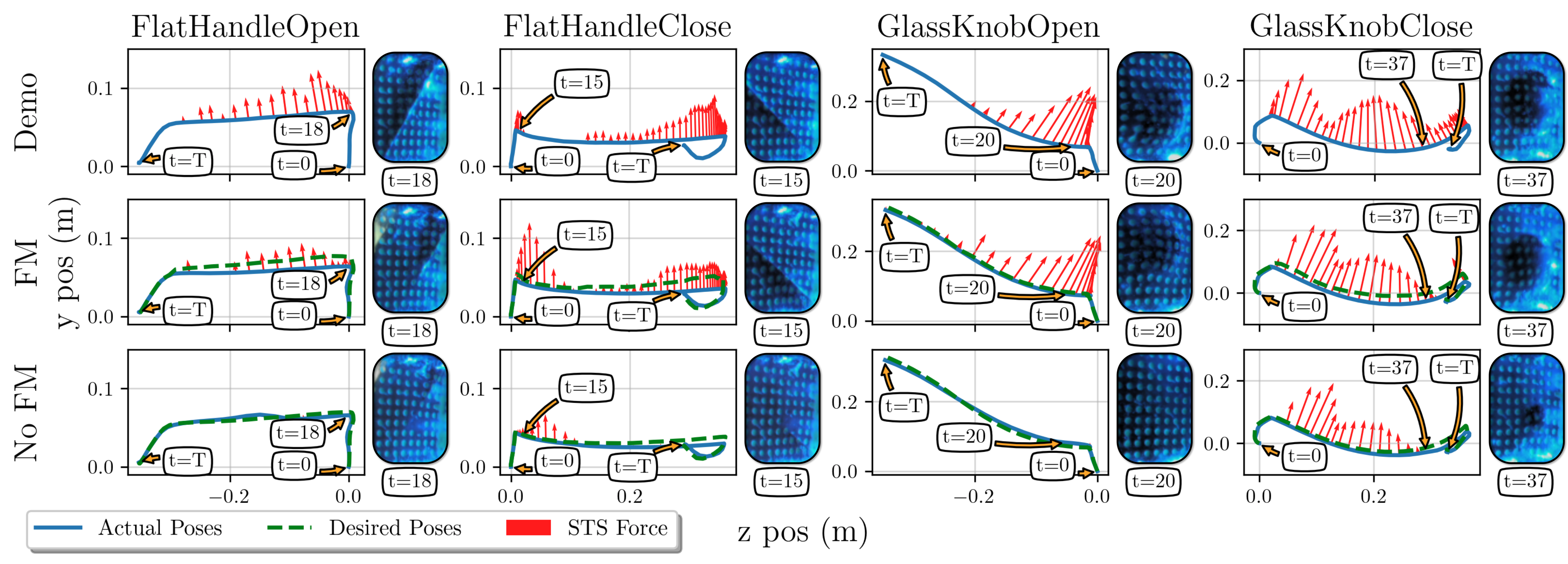}
    \caption{
    \hl{Top: a raw demonstration trajectory $\tau_{x,\text{raw}}$ (blue) as well as demonstration forces $\tau_{\Tilde{f},\text{raw}}$ (red).
    Middle: a new set of desired poses that incorporate force matching $\tau^d_{x,\text{rep}}$ (green), the new set of replayed poses $\tau_{x,\text{rep}}$ given $\tau^d_{x,\text{rep}}$ (blue), and the actual forces with the modified trajectory $\tau_{\Tilde{f},\text{rep}}$ (red).
    Bottom: a set of desired and actual poses, along with resulting forces, that use $\tau_{x,\text{raw}}$ directly (i.e., \textit{without} force matching), while ignoring $\tau_{\Tilde{f},\text{raw}}$, to generate a replay (green, blue, and red, respectively).
    }}
    \label{fig:fm_with_without}
    \vspace{-.3cm}
\end{figure*}

As detailed in \cref{sec:method_force_measuing}, a short calibration procedure allows us to solve for the parameters in \cref{eqn:new_desired_poses_actual}.
This procedure takes about seven minutes in practice, and is fully autonomous.
We use our glass knob as the calibration object, and move the desired pose of the end-effector 3.5 cm towards the glass knob while moving at 1 mm increments, with each trajectory starting between 1 mm and 3 mm away from the knob.
We also generate shear by moving 1 cm laterally, and torque by moving 1 rad rotationally, after already having made initial contact with the knob.

\subsection{Imitation Learning and Training Parameters}
\label{sec:il_training_parameters}

\begin{figure}[!b]
    \vspace{-.3cm}
    \centering
    \includegraphics[width=.9\columnwidth]{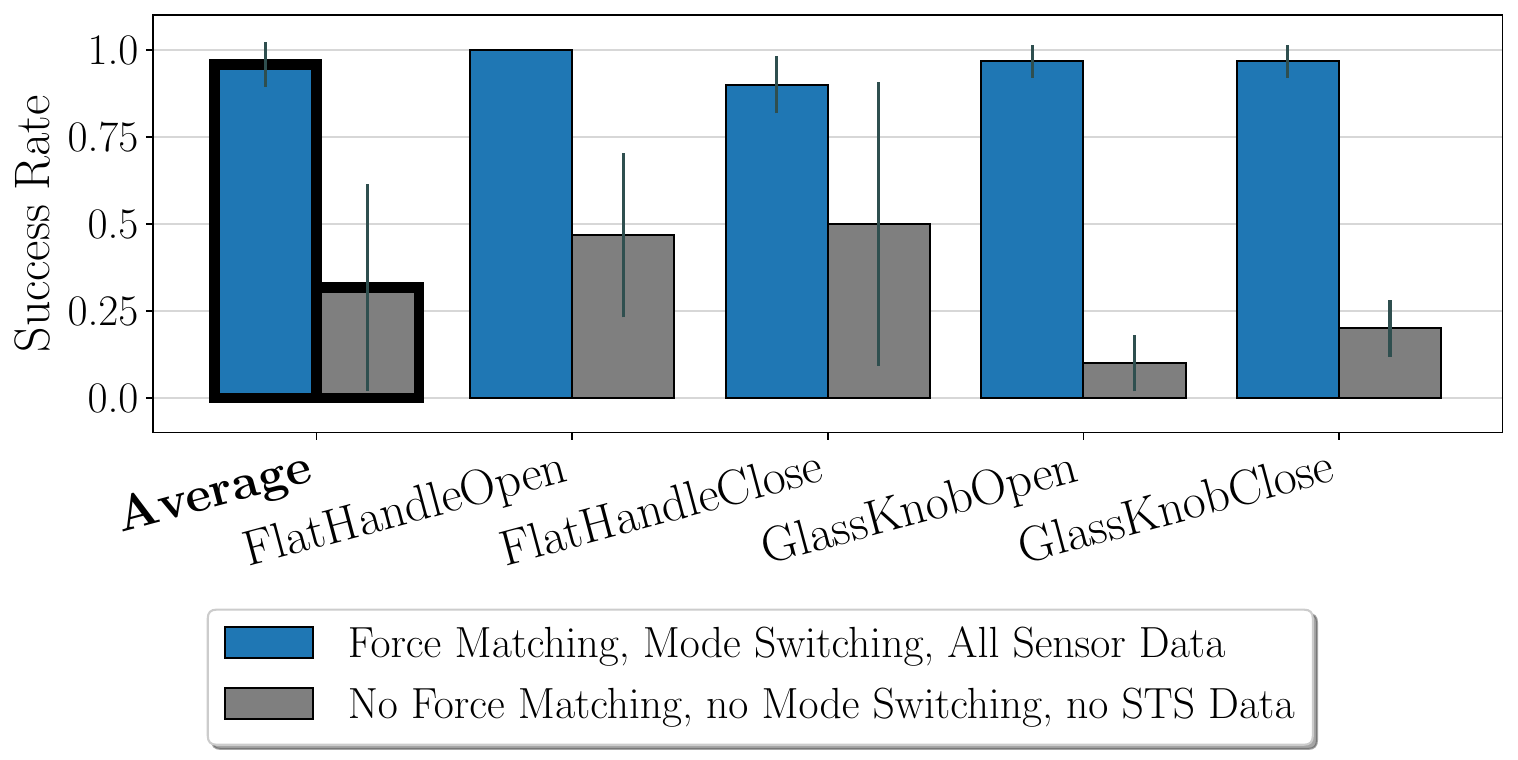}
    \caption{Performance results \hl{with and without each of the three novel additions presented in this work:} force matching, STS mode switching, and STS as a policy input.
    There is a clear benefit across all tasks, with greater benefit for each of the \texttt{GlassKnob} tasks.
    For this and the following figures, the average across all tasks is shown in bold on the left and black lines indicate standard deviation of seeds.}
    \label{fig:perf_system}
\end{figure}

Kinesthetic teaching data  are collected at \hl{10 Hz} once the robot has exceeded a minimum initial movement threshold \hl{of 0.5 mm.}
After a kinesthetic demonstration is completed, the \hl{robot} is reset to the same initial (global) pose, and the demonstration is replayed \hl{with or without force matching, and with or without the human providing a mode-switch label.}
\hl{We} collected 20 raw kinesthetic teaching demonstrations with a human expert \hl{for each task (80 total).}
With these demonstrations, we complete replay trajectories in a variety of configurations (described in \cref{sec:force_matching_perf,sec:policy_mode_switching_perf,sec:exp_data_scaling}, depending on which algorithm \hl{is being tested.}
For every configuration, we train policies with three different random seeds, and complete 10 test episodes \hl{per} seed.

We trained our policies in PyTorch with the Adam optimizer and a learning rate of 0.0003, halving the learning rate halfway through training.
We use a ResNet-18 \cite{heDeepResidualLearning2016} architecture pretrained on ImageNet and ending with spatial soft-argmax \cite{levineEndtoendTrainingDeep2016} for image data, and a small fully connected network for relative pose data.
Features from each modality are concatenated and passed through another small fully connected network before outputting our seven-dimensional action: relative position change, orientation change as a rotation vector, and STS mode.
All layers use ReLU non-linearities.
We train each policy for 20k gradient steps, using weight decay of 0.1 to avoid overfitting, as this has been shown to improve behavioural cloning results compared to early stopping \cite{mandlekarWhatMattersLearning2021,ablettLearningGuidedPlay2023}.

\subsection{System Performance}
\label{sec:system_perf}

We evaluate the benefits of collectively including force matching, mode switching as a policy output, and STS data as a policy input by training policies with all three of these additions, as well as policies with none of them, and comparing their success rates.
\hl{As stated in \cref{sec:il_training_parameters}, we train three seeds per task with 10 test episodes per seed, that is, 30 episodes for each task and configuration, and 120 total test episodes per configuration.}
The comparison is visualized for each individual task, and with an overall average across all tasks, in \cref{fig:perf_system}.
\hl{Including the three additions results in a 64.2\% average absolute across-task performance gain over the baseline where none are included, demonstrating a clear benefit.}
There is a task-related correlation also, where the average improvement for the \texttt{FlatHandle} tasks is 46.7\%, while the average improvement for \texttt{GlassKnob} tasks is 81.7\%.
The remaining sections ablate each of these additions individually to better understand each component's contribution.

\subsection{Force Matching Performance}
\label{sec:force_matching_perf}

\begin{figure}[b]
    \vspace{-.4cm}
    \centering
    \includegraphics[width=.9\columnwidth]{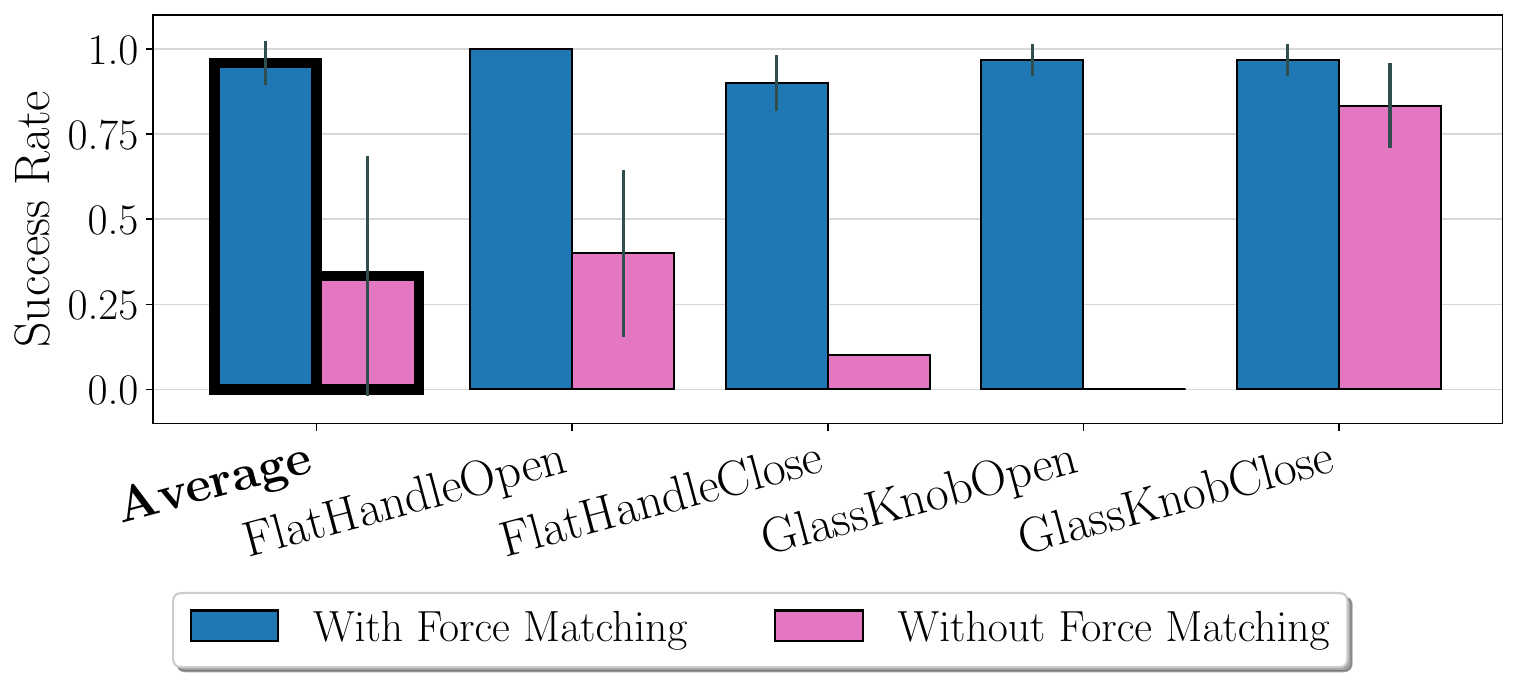}
    \caption{Performance results showing the effect of excluding force matching, while keeping mode switching and the STS as a policy input in both cases. 
    Force matching improves performance in all tasks, with particularly large gains for \texttt{FlatHandleClose} and \texttt{GlassKnobOpen}.
    }
    \label{fig:perf_fm}
\end{figure}

To evaluate the benefit \hl{of force matching,} we complete replays of our raw expert datasets both with and without force matching and compare their performance.
To isolate the benefits of force matching, and reduce the effect of including or excluding policy mode switching, we include policy mode switching in both configurations.
We also include all three sensing modalities (wrist-mounted eye-in-hand RGB images, STS images, and relative positions) in \hl{both.}
The results of these tests are shown in \cref{fig:perf_fm}.

\hl{The average absolute across-task performance increase with force matching is 62.5\%, although the performance gain is greater for some tasks than for others.}
A notable case is the discrepancy between gains for the two \texttt{GlassKnob} tasks: \hl{the \texttt{GlassKnobOpen} policy always fails without force matching, while the \texttt{GlassKnobClose} policy sees only a slight benefit from force matching.}
This can partially be explained by the pose angle for much of the demonstration.
This can partially be explained by the pose angle for much of the demonstration.
For the \texttt{GlassKnobOpen} task, after contact is made with the knob, the door is almost exclusively pulled via a shear force.
Poor initial contact caused by too little force \hl{often} results in failure (see third column, third row of \cref{fig:fm_with_without}).
In contrast, for the \texttt{GlassKnobClose} task, the angle of the robot finger relative to the knob ensures that normal force \hl{is still applied} throughout much of the \hl{trajectory.}

\hl{The force-matching results in \cref{fig:perf_fm} may appear contradictory to those from \cref{fig:perf_system}.
In \cref{fig:perf_fm}, the use of STS data without force matching actually performs \textit{worse}, for 
\texttt{FlatHandleClose} and \texttt{GlassKnobOpen}, than excluding STS data and force matching (as shown in \cref{fig:perf_system}).
We hypothesize that this is due to causal mismatch \cite{dehaanCausalConfusionImitation2019}: the highly suboptimal demonstration data generated without force matching may cause the policy to learn to switch the STS modality without firm contact.
This then causes the policy to initiate the ``open" or ``close" phase (as opposed to the ``reach" phase) of the task too early.
We leave further investigation of this result to future work.}

\subsubsection{Force Matching Trajectory Comparison}
\label{sec:force_matching_trajectory_comparison}

While our policy learning experiments in \cref{sec:force_matching_perf} implicitly illustrate the value of force matching, \cref{fig:fm_with_without} shows specific examples of how trajectories change with and without force matching.
We label the initial timestep $t=0$, final timestep $t=T$, and a single representative timestep for each trajectory, also including a cropped STS image at the representative timestep for each trajectory.
Note that the No FM desired poses (bottom, green) are the same as the true demo poses (top, blue), \hl{while the difference between the desired poses and actual poses in the FM row lead to increased STS force (see \cref{eqn:new_desired_poses_actual}).}
In all four cases, the No FM replay causes the end-effector to slip off of the handle or knob, leading to failure. 

\begin{figure}[t]
    \centering
    \includegraphics[width=.9\columnwidth]{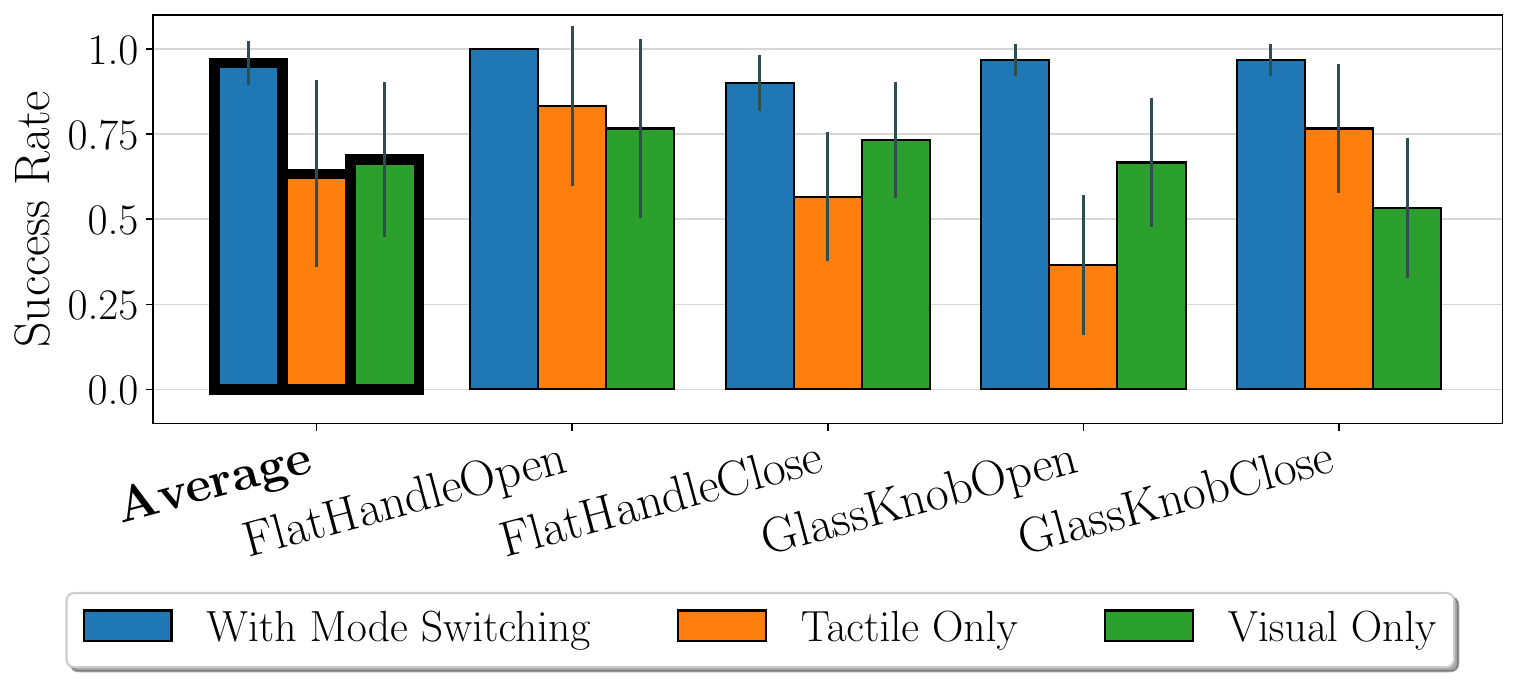}
    \caption{Performance results comparing the use of mode switching with setting the STS sensor to visual-only and tactile-only. 
    Mode switching provides a clear benefit over keeping the sensor in a single mode, but there is no obvious pattern between whether a tactile-only or visual-only sensor would be preferred without mode switching.}
    \label{fig:perf_ms}
    \vspace{-.3cm}
\end{figure}

\begin{figure}[b]
    \vspace{-.3cm}
    \centering
    \includegraphics[width=\columnwidth]{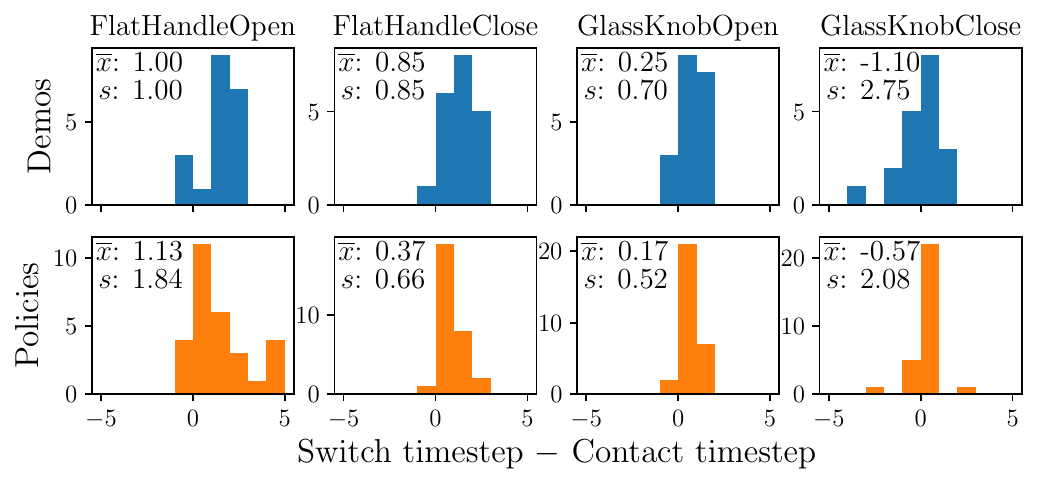}
    \caption{Histograms of the difference between mode switch action timestep and the true first contact timestep, for both our demonstrations and our learned policies with force matching, mode switching, and all sensing modalities.}
    \label{fig:mode_contact_diff_histograms}
\end{figure}

We conclude that force matching generates $\tau_{x,\text{rep}}$ and $\tau_{\Tilde{f},\text{rep}}$ that better match $\tau_{x,\text{raw}}$ and $\tau_{\Tilde{f},\text{raw}}$.
While $\tau_{\Tilde{f},\text{rep}}$ occasionally has mismatches with $\tau_{\Tilde{f},\text{raw}}$, most of these errors can be attributed to a combination of sensor reading errors \hl{(see noise in bottom of \cref{fig:calibration_normal_example})} and control \hl{errors.}

\subsection{Policy STS Mode Switching Performance}
\label{sec:policy_mode_switching_perf}

\begin{figure*}[t]
    \centering
    \includegraphics[width=.9\textwidth]{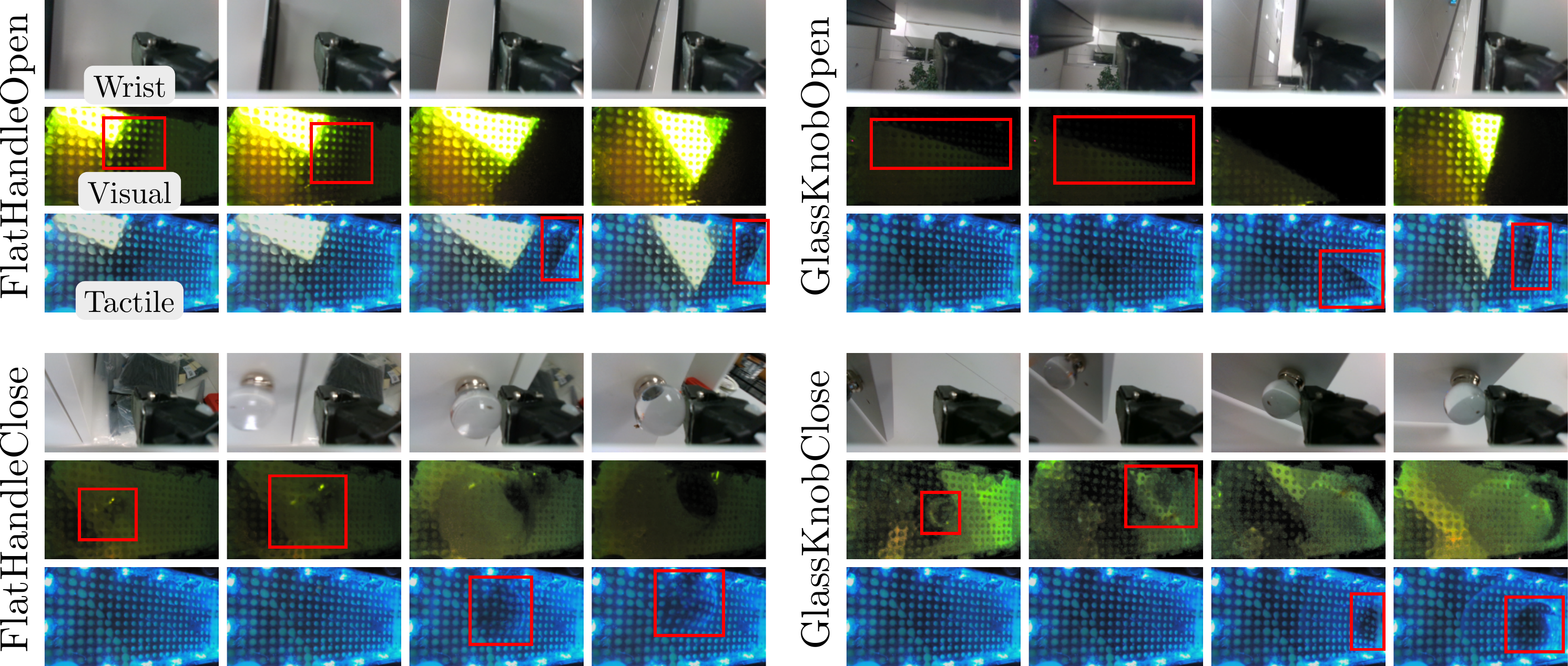}
    \caption{\hl{Partial trajectories from the same replay in visual-only and tactile-only for all four tasks.}
    Red boxes highlight parts of the scene (knob, handle) that are significantly clearer \hl{in the respective sensor mode.}
    The corresponding wrist camera \hl{images are provided on the top rows} for reference.}
    \label{fig:visual_tactile_only}
    \vspace{-.3cm}
\end{figure*}

\hl{To better understand how mode switching can benefit performance on a real door-opening task, we collect replays of demonstrations both with and without mode switching.}
Specifically, we consider three configurations: mode switching enabled, tactile-only (the sensor lights are always on), and visual-only (the sensor lights are always off).
Given the benefit of including force matching (as shown in \cref{sec:force_matching_perf}) and to isolate the benefits of mode switching, we included force matching in all three configurations.
\hl{We also include all three sensing modalities.} 
Our results are shown in \cref{fig:perf_ms}, using the same training and testing parameters as described in \cref{sec:il_training_parameters}.

Averaging the success rates of the visual-only and tactile-only results together, the use of mode-switching results in an a task-average performance gain of 30.4\%.
As with force matching, there are some task-specific patterns.
\hl{Task-specific correlations are similar to \cref{sec:system_perf}: the performance gain for \texttt{GlassKnob} tasks is 38.3\%, while the performance gain for \texttt{FlatHandle} tasks is only 22.5\%.}
This is perhaps unsurprising, given that the handle \hl{requires less precision to maintain contact.}
\hl{Qualitatively, \cref{fig:visual_tactile_only} shows the value of including sensor mode switching: tactile-only data provides no information before contact, while visual-only data provides poor tactile information during contact.}

\subsubsection{Mode Switching Timing Analysis}
\label{sec:mode_switching_timing_analysis}

To \hl{evaluate} the quality of our learned mode switching policy output, we completed an analysis of the timing of the learned mode switch action compared with the expert labels.
An expert demonstrator may have a preference for having the sensor switch modes slightly before or after contact occurs, but for this analysis, we will consider that an optimal switch occurs at the moment of contact.

For each dataset with force matching, mode switching, and all three sensing modalities, we manually annotated each episode with the timestep at which contact was made between the handle/knob and the surface of the STS, and compared that to the timestep that the demonstrator provided a mode switch action label.
The histograms of these timestep differences are shown in the top row of \cref{fig:mode_contact_diff_histograms}.
While there are certain task-specific patterns, such as a slightly greater timestep difference average for the handle tasks, the clearest pattern is that the mode switch label usually occurs within one timestep (0.1s) of contact being made.
The bottom row shows the same analysis, but for autonomous policies.
With the exception of a few outliers (e.g., for \texttt{FlatHandleOpen}), the policies have converged to be close to an average of zero timesteps between contact and mode switching (i.e., smoothing out the reactive/predictive timing errors from the expert dataset).
The outliers in Flat Handle are \hl{most likely} due to a causal mismatch, where the policy learns to switch modes based on when the arm starts opening the handle, instead of at the moment of contact \cite{dehaanCausalConfusionImitation2019}.

\subsection{Observation Space Study}
\label{sec:exp_obs_study}

\begin{figure}[b]
    \vspace{-.3cm}
    \centering
    \includegraphics[width=.9\columnwidth]{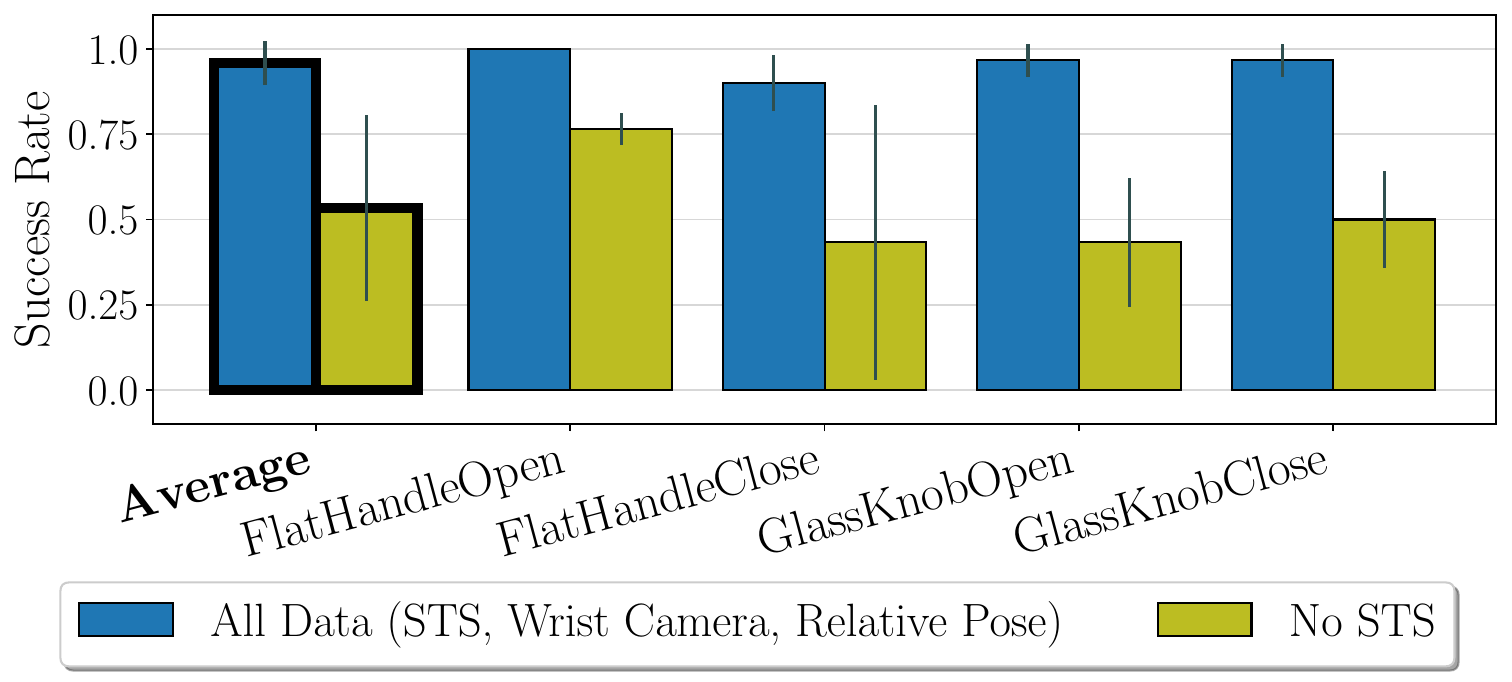}
    \caption{Performance when \hl{STS data are excluded} as a policy input.
    Force matching is still included in this case.
    The data from the STS clearly provide substantial benefit for learned policies.
    }
    \label{fig:wrist_vs_no_wrist}
\end{figure}

The \hl{second} experimental question we hope to answer is whether our door manipulation tasks can be complete with a wrist-mounted eye-in-hand camera alone, and whether the inclusion of STS data improves performance.
\hl{To complete these tests, we do not collect any new human demonstration data or any new replay data, instead we selectively exclude some observations during training.} %
\hl{All policies are trained with force matching enabled, and policies trained with STS data include mode switching.}
For policies that exclude STS data, we set the sensor to visual-only mode, since the wrist camera \hl{captures STS lighting changes that might provide a contact cue.}

Results from these tests are shown in \cref{fig:wrist_vs_no_wrist}.
\hl{Adding the STS sensor as an input yields in an average across-task increase in performance of 42.5\%.}
It is worth reiterating that this performance gain only corresponds to the case where STS data is excluded as an input into the policy; the STS itself is still used indirectly for these policies through the use of force matching.
\hl{Elaborating, for this baseline, the STS sensor is used exclusively in tactile mode and could be replaced with a regular visuotactile sensor or a finger-mounted force-torque sensor.}
Without force matching, the performance gain over the baseline is 64.2\% (as shown in \cref{sec:system_perf}), indicating that force matching alone provides some benefit, but that the STS as a policy input is still quite \hl{valuable.}
\hl{The average performance gain of 50\% for the \texttt{GlassKnob} policies is again higher than 35\% for the \texttt{FlatHandle} policies, indicating once more that the utility of the STS as a policy input is partially task-dependent.}

\subsubsection{Contribution of Relative Pose Input}
\label{sec:exp_relative_pose_input}

\begin{figure}[b]
    \vspace{-.3cm}
    \centering
    \includegraphics[width=.9\columnwidth]{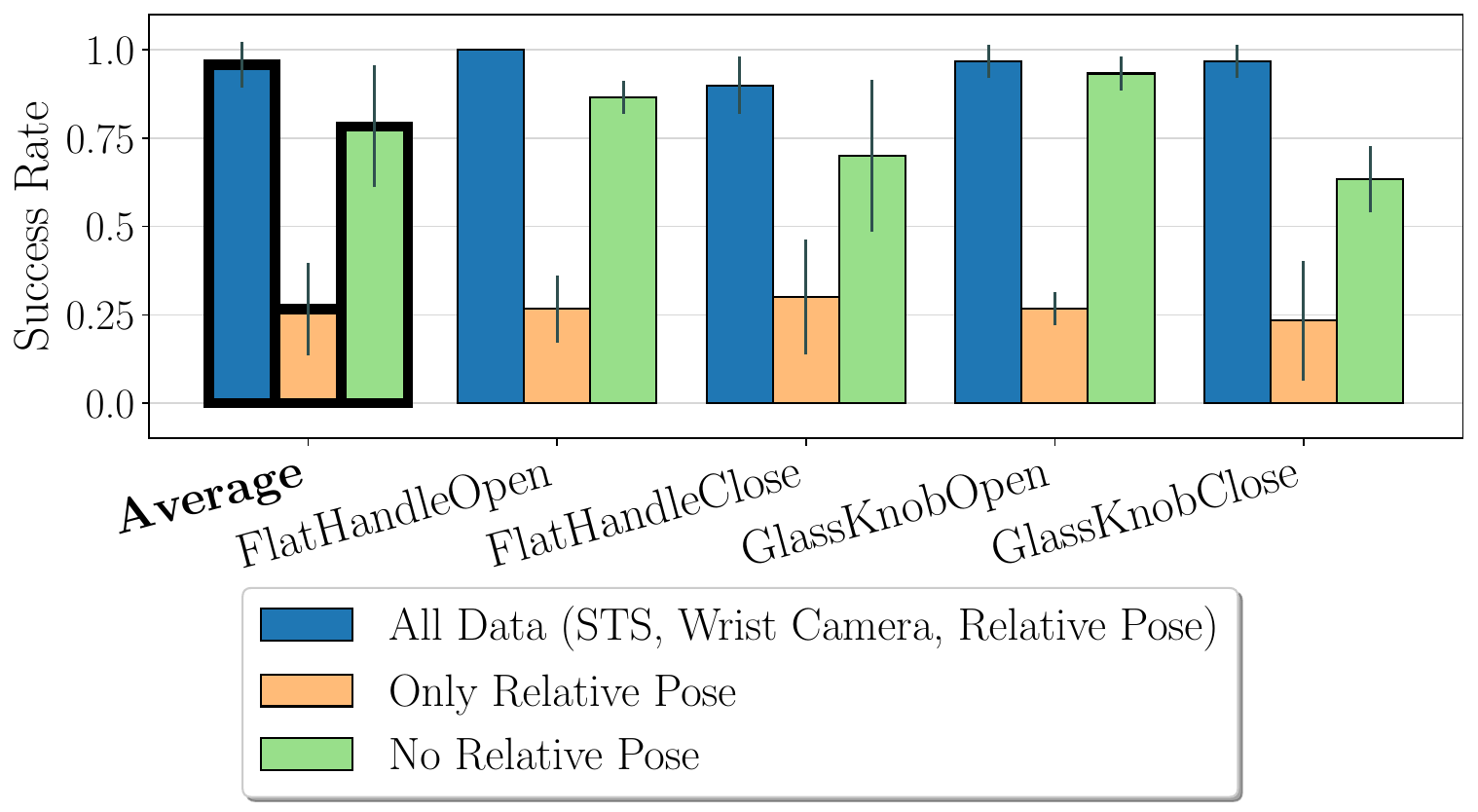}
    \caption{Performance results to illustrate the contribution of relative pose as a policy input. 
    Relative pose alone is not capable of solving the tasks, but its inclusion along with each of our sensors does have a positive effect on performance.
    }
    \label{fig:rel_pose_vs_no_rel_pose}
\end{figure}

\hl{To better understand the value of relative poses as inputs, we train policies with and without relative pose, as well as policies with \textit{only} relative pose as an input (and no other changes).}
\hl{Results for training and testing in} these three configurations are shown in \cref{fig:rel_pose_vs_no_rel_pose}.
\hl{The relatively high performance} of policies that exclude relative pose shows that the visual data alone \hl{often provides enough information to solve the task.}
\hl{However, including relative pose still yields an average across-task policy improvement of 17.5\%.}

\subsection{STS-Only Policy Performance}
\label{exp:sts_only_study}

\begin{figure}[t]
    \centering
    \includegraphics[width=.9\columnwidth]{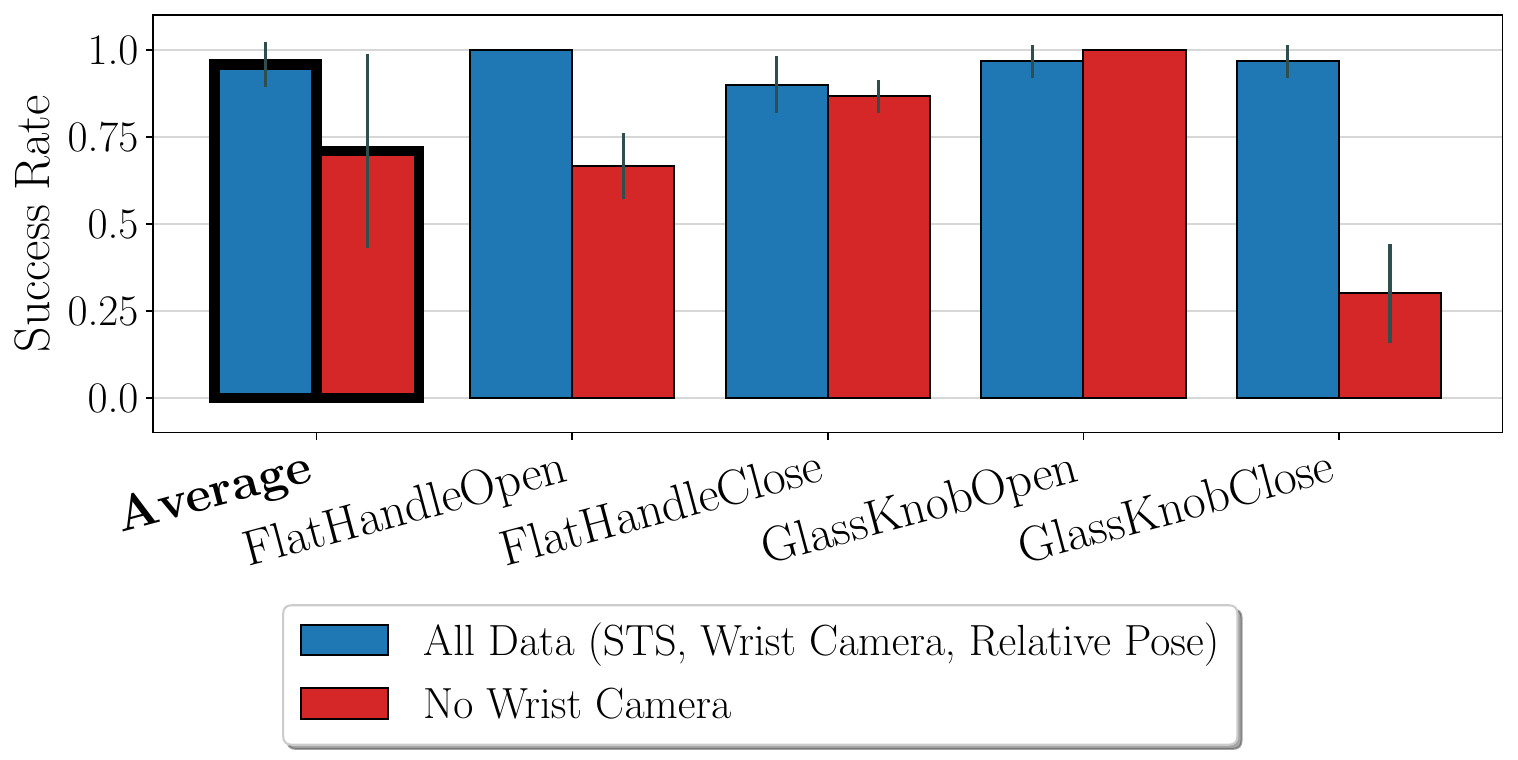}
    \caption{A performance comparison of the effect of excluding the eye-in-hand wrist-mounted RGB camera data while keeping both force matching and mode switching enabled.
    \hl{Performance deteriorates slightly in \texttt{FlatHandleOpen} and dramatically in \texttt{GlassKnobClose}, but an STS-only policy does perform adequately in certain cases.}
    }
    \label{fig:perf_sts_only}
    \vspace{-.3cm}
\end{figure}

\hl{Our third experimental question concerns whether an STS camera alone provides enough feedback to complete these challenging door manipulation tasks.}
\hl{Prior work has shown that simply adding more sensor data} to learned models does not always lead to improved performance \cite{limoyoHeteroscedasticUncertaintyRobust2020, hansenVisuotactileRLLearningMultimodal2022, mandlekarWhatMattersLearning2021}.
In this section, we train \hl{and test} several policies without using data from the wrist-mounted eye-in-hand \hl{camera.}

The results in \cref{fig:perf_sts_only} show an average across-task absolute performance increase of 25.0\% with the wrist camera data.
\hl{The amount of improvement varies significantly by task, however, with no improvement for \texttt{FlatHandleClose} and \texttt{GlassKnobOpen}, but a dramatic improvement for \texttt{GlassKnobClose}.}
\hl{This is at least partially because the initial reach in \texttt{GlassKnobClose} is more difficult than for any of the other tasks, and the wrist camera provides a clear view of the approach.}
\hl{Excluding the wrist camera data (\cref{fig:perf_sts_only}, red) from policy inputs results in a 17.5\% higher success rate than excluding the STS data (\cref{fig:wrist_vs_no_wrist}, light green).}
This finding \hl{may indicate} that contact-rich door tasks benefit more from a multimodal STS sensor alone than from a wrist camera alone.

\subsubsection{Mode Switching Performance (STS-Only)}
\label{sec:policy_mode_switching_perf_sts_only}

\begin{figure}[b]
    \vspace{-.3cm}
    \centering
    \includegraphics[width=.9\columnwidth]{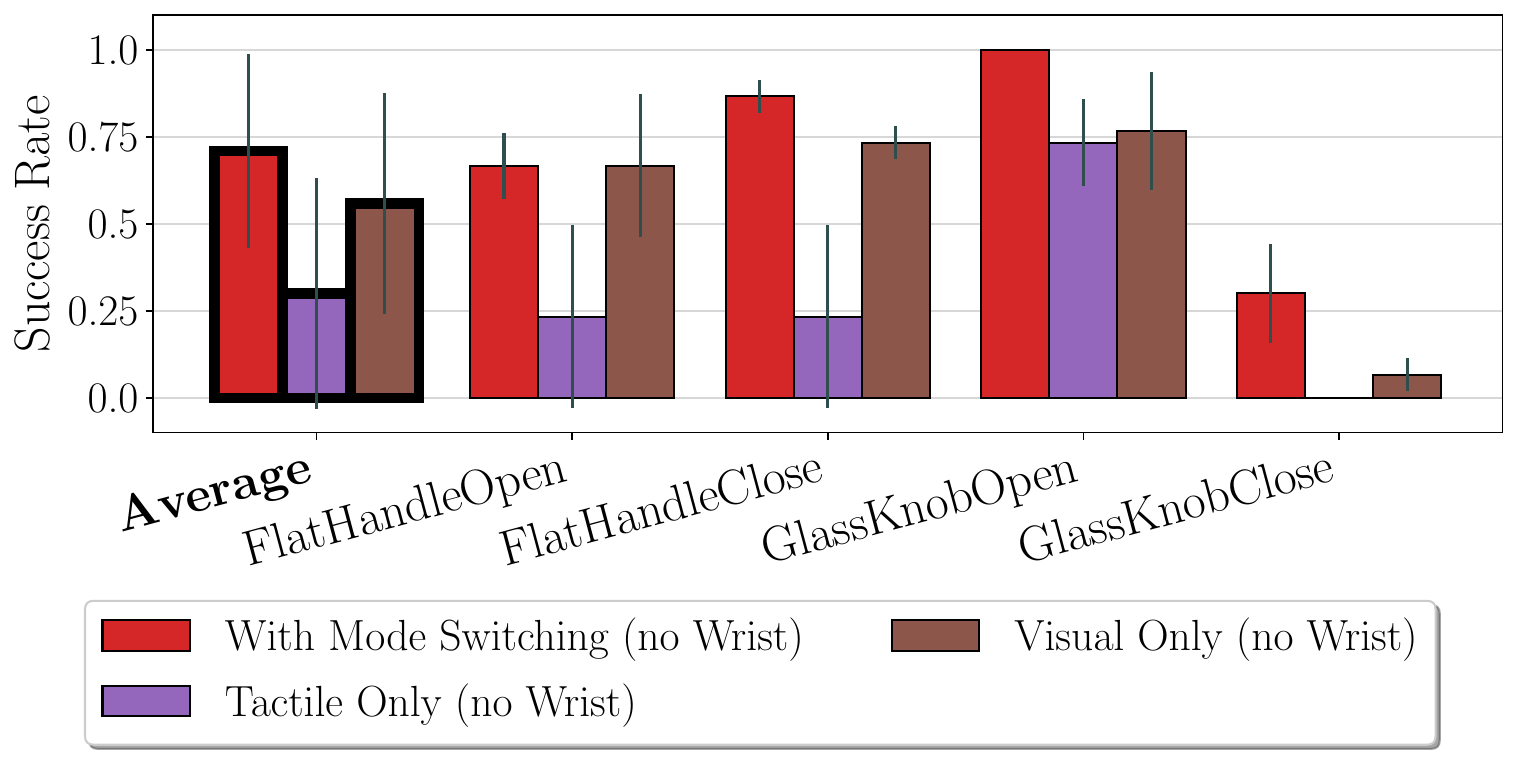}
    \caption{An STS-only variant of the mode switching performance results shown previously in \cref{fig:perf_ms}.
    A similar pattern emerges, highlighting the value of mode switching. 
    In the STS-only case, tactile-only policies are shown to be significantly poorer than visual-only policies.}
    \label{fig:perf_ms_sts_only}
\end{figure}

\hl{We also evaluate the benefits of our policy mode switching output, replicating the experiments in \cref{sec:policy_mode_switching_perf}, but excluding the wrist camera data.}
\hl{Specifically, we train and test policies using STS data and in three configurations: with mode switching enabled, with the STS mode set to visual-only, and with the mode set to tactile-only.}
As in \cref{sec:policy_mode_switching_perf}, \hl{we include both force matching and relative poses.}
\hl{Results are shown in \cref{fig:perf_ms_sts_only}.
The average across-task performance gain, for visual-only and tactile-only combined, is 27.9\%. 
This is comparable to the increase of 30.4\% shown in \cref{sec:policy_mode_switching_perf} when wrist camera data is included.}

\begin{figure}[t]
    \centering
    \includegraphics[width=\columnwidth]{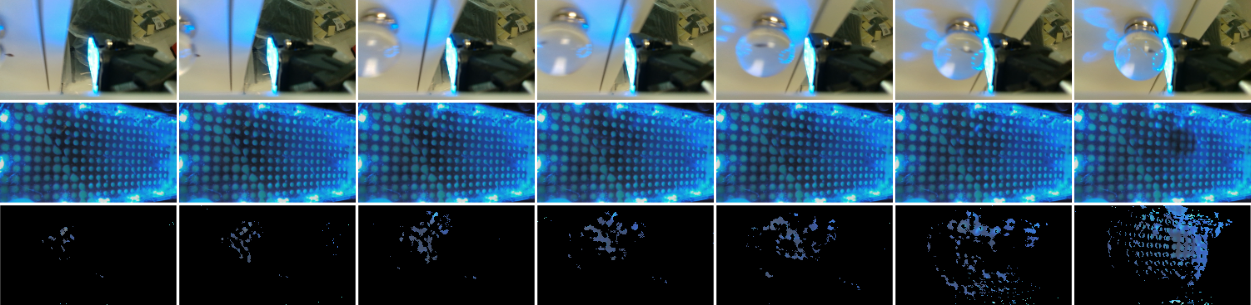}
    \caption{A short sub-trajectory of a \texttt{GlassKnobOpen} run showing the reaching phase with the STS sensor set to tactile-only \hl{mode}.
    The top row shows the wrist camera view for reference, the middle row shows the (tactile) STS images, and the bottom row shows pixels that have changed since the previous timestep. 
    The glass knob shows up faintly, explaining the surprisingly good performance of STS-only tactile-only policies in \texttt{GlassKnobOpen} from \cref{fig:perf_ms_sts_only}.
    }
    \label{fig:glass_knob_tactile_only_example}
    \vspace{-.3cm}
\end{figure}

\hl{One striking difference is the performance increase from mode switching over tactile-only policies of 40.8\%, compared to only a 15.0\% increase over visual-only policies.}
\hl{These tasks require reaching to make adequate contact with the handle or knob, and with both visual STS data and eye-in-hand wrist data absent, the tasks become much more difficult.}
\hl{Tactile-only performance matches visual-only performance for \texttt{GlassKnobOpen}, which may be because, even in tactile mode, the glass knob faintly shows up in the STS sensor images (see \cref{fig:glass_knob_tactile_only_example}).}
\hl{However, even for \texttt{GlassKnobOpen}, mode switching is clearly the optimal configuration in which to use the STS sensor.}

\subsection{Expert Data Scaling}
\label{sec:exp_data_scaling}

\begin{figure}[b]
    \vspace{-.3cm}
    \centering
    \includegraphics[width=.95\columnwidth]{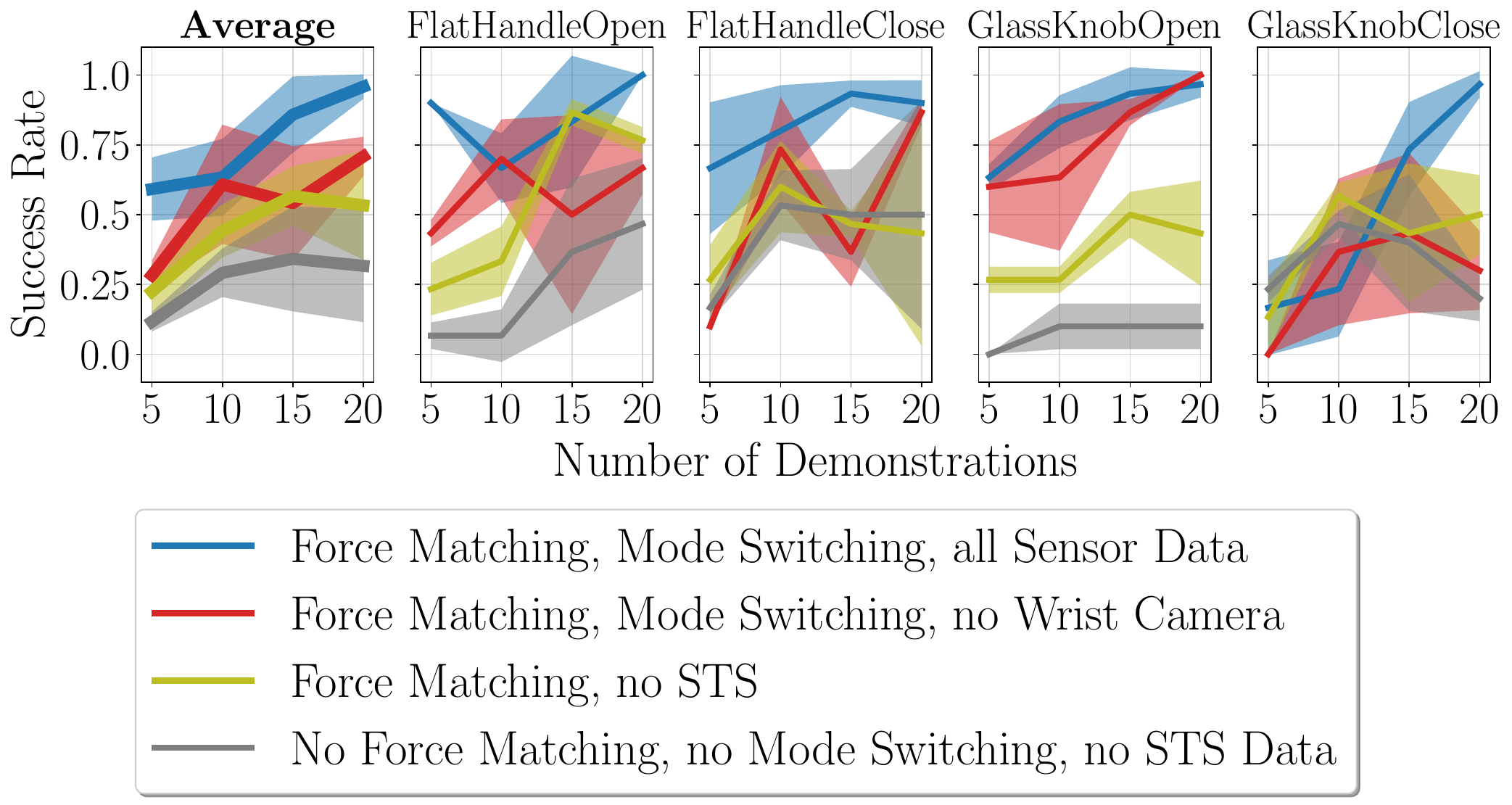}
    \caption{Performance with varying amounts of expert data for representative configurations from our prior experiments.
    The shading indicates one standard deviation.
    All modalities show increasing performance with increasing dataset sizes, but the two modalities without STS data reach a plateau at 15 demonstrations.
    }
    \label{fig:perf_data}
\end{figure}

In our final set of experiments, we choose representative configurations from the preceding sections and train and test policies with 5, 10, and 15 expert trajectories (compared with 20 for all of our previous experiments).
\cref{fig:perf_data} shows the results of experiments with \hl{less} training data for these configurations.
Most methods improve with more data, but there are exceptions.
For example, both STS-free methods do not improve significantly for \hl{\texttt{FlatHandleClose}, \texttt{GlassKnobOpen}, or \texttt{GlassKnobClose},} indicating that the eye-in-hand wrist camera \hl{does not provide sufficient} information to complete the task alone, regardless of data quantity.
Conversely, eye-in-finger visual data, as well as tactile data, \hl{are sufficient for these three tasks,} so their performance scales up with increasing data.
A surprising finding is that for \texttt{FlatHandleOpen}, \texttt{FlatHandleClose}, and \texttt{GlassKnobOpen}, performance is quite good with force matching, mode switching, \hl{and across all sensing modalities}, even with only five demonstrations.
Performance on \hl{the most difficult task, \texttt{GlassKnobClose}, clearly increases with more data, but the STS data alone are not sufficient to complete this task.}
\hl{Results for the two configurations that exclude STS input data plateau at 15 demonstrations, indicating that these tasks may be simply not possible to complete without the STS, even with an increasing amount of data.}

\hl{\subsection{Decoupling Human Forces from Contact Forces}
\label{sec:decoupling_human_forces}

As described in \cref{sec:method_force_measuing}, common approaches to sensing end-effector force-torque, such as wrist-mounted force-torque sensors or joint-torque sensors with dynamics modelling, cause recorded wrenches $\tau_{f,\text{raw}}$ to be corrupted by the demonstrator's own force against the robot.
To avoid this corruption, a sensor must be mounted at a point on the end-effector where human-applied forces are ignored during demonstrations, such as on the finger of a gripper.

\begin{figure}[t]
    \centering
    \includegraphics[width=\columnwidth]{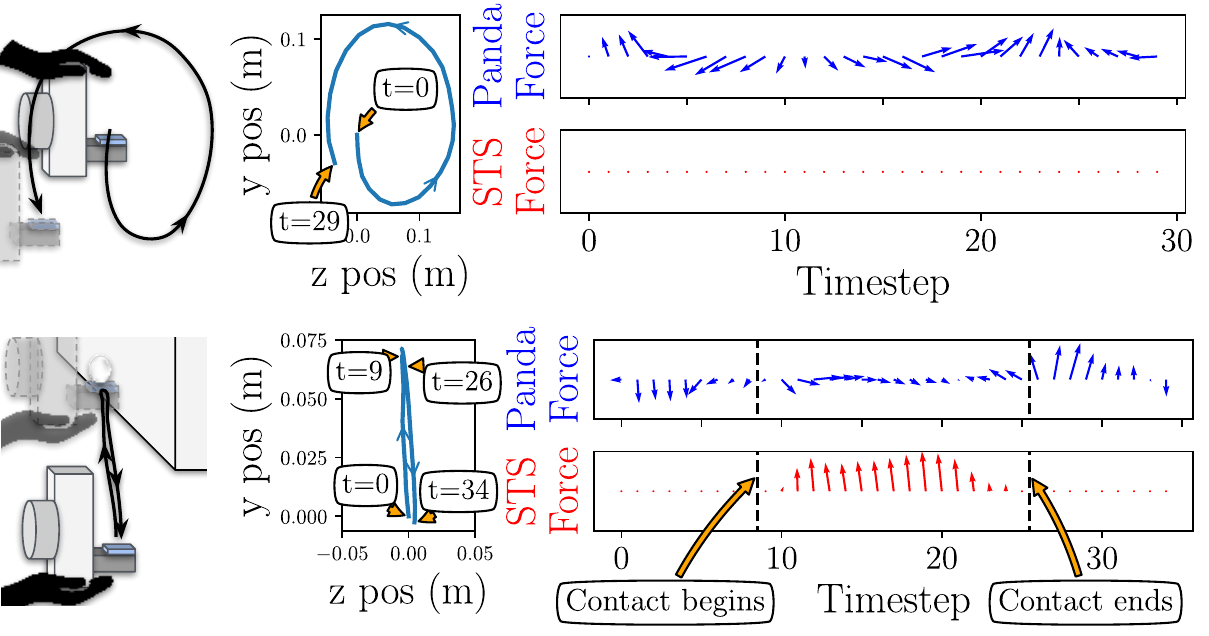}
    \caption{Example expert demonstrations showing measured forces (collapsed to 2-DOF) with Panda (the robot) joint-torque sensors and dynamics modelling (blue) and our own sensor (red).
    The top trajectory corresponds to pushing the end-effector in a circle without environmental contact, and the bottom trajectory corresponds to pushing the end-effector towards, against, and away from the door knob while fixed.
    Panda force readings are corrupted by human-robot interaction forces.
    }
    \label{fig:panda_ft_comp}
    \vspace{-.3cm}
\end{figure}

In \cref{fig:panda_ft_comp}, we validate the presence of this corruption, and its resolution through the use of finger-mounted wrench sensing.
The STS sensor correctly shows no readings in the circular trajectory, where there is no contact, and shows increasing and then decreasing force during contact in the trajectory where it is pushed against the knob.
Conversely, the Panda force readings are primarily in response to the human pushing force against the robot, making it difficult to isolate end-effector to environment forces.
}

\section{Limitations}
\label{sec:limitations}

In this section, we discuss some limitations of our work.
Although we used a single sensor for all of our experiments, the gel-based contact surface of the STS physically degrades over time.
This limitation partially motivates the use of learned models for gel-based sensors; if sensor data changes marginally due to degradation, a practitioner can simply add more data to the dataset and retrain the policy.
As well, the STS incorporates a standard camera, so policies trained on its data are susceptible to the same problems as other visual data paired with neural networks, such as overfitting to specific lighting conditions.
Finally, our force-matched replayed demonstrations still occasionally fail because both our method for measuring forces as well as our Cartesian impedance controller can suffer from accuracy issues, but both of these limitations can be improved with further tuning.

\section{Conclusion}
\label{sec:conclusion}

\hl{In this paper, we presented a robotic imitation learning system that leverages a see-through-your-skin (STS) visuotactile sensor as both a measurement device to improve demonstration quality and a multimodal source of raw data to learn from.}
Our first contribution was a \hl{method to use the STS in its tactile mode as a force-torque sensor to improve demonstration quality through force-matched replays that better recreate the demonstrator's force profile.}
Our second contribution was \hl{a learned approach to STS mode switching, in which a policy output is added to switch the sensor mode on the fly, and labels are provided by the demonstrator during demonstration replays.}
Our final contribution was an observational study in which we compared and contrasted the value of the STS visuotactile data (in mode-switching, visual-only, and tactile-only configurations) \hl{with the use of an eye-in-hand wrist-mounted camera on four challenging, contact-rich door opening and closing tasks on a real manipulator.}
Potential directions for future work include improving the accuracy of our \hl{force sensing approach and adding a means for self-improvement by automatically detecting execution failures.}

\section*{Acknowledgment}
This work was supported in part by the Natural Sciences and Engineering Research Council of Canada (NSERC).
Jonathan Kelly gratefully acknowledges support from the Canada Research Chairs program.

\bibliographystyle{IEEEtran}
\bibliography{sts-il}  %

\begin{thebibliography}{10}
\providecommand{\url}[1]{#1}
\csname url@rmstyle\endcsname
\providecommand{\newblock}{\relax}
\providecommand{\bibinfo}[2]{#2}
\providecommand\BIBentrySTDinterwordspacing{\spaceskip=0pt\relax}
\providecommand\BIBentryALTinterwordstretchfactor{4}
\providecommand\BIBentryALTinterwordspacing{\spaceskip=\fontdimen2\font plus
\BIBentryALTinterwordstretchfactor\fontdimen3\font minus \fontdimen4\font\relax}
\providecommand\BIBforeignlanguage[2]{{%
\expandafter\ifx\csname l@#1\endcsname\relax
\typeout{** WARNING: IEEEtran.bst: No hyphenation pattern has been}%
\typeout{** loaded for the language `#1'. Using the pattern for}%
\typeout{** the default language instead.}%
\else
\language=\csname l@#1\endcsname
\fi
#2}}

\bibitem{chiRecentProgressTechnologies2018}
C.~Chi, X.~Sun, N.~Xue, T.~Li, and C.~Liu, ``Recent {{Progress}} in {{Technologies}} for {{Tactile Sensors}},'' \emph{Sensors}, vol.~18, no.~4, p. 948, Apr. 2018.

\bibitem{yuanGelSightHighResolutionRobot2017}
W.~Yuan, S.~Dong, and E.~H. Adelson, ``{{GelSight}}: {{High-Resolution Robot Tactile Sensors}} for {{Estimating Geometry}} and {{Force}},'' \emph{Sensors}, vol.~17, no.~12, p. 2762, Dec. 2017.

\bibitem{padmanabhaOmniTactMultiDirectionalHighResolution2020}
A.~Padmanabha, F.~Ebert, S.~Tian, R.~Calandra, C.~Finn, and S.~Levine, ``{{OmniTact}}: {{A Multi-Directional High-Resolution Touch Sensor}},'' in \emph{2020 {{IEEE International Conference}} on {{Robotics}} and {{Automation}} ({{ICRA}})}, May 2020, pp. 618--624.

\bibitem{maDenseTactileForce2019}
D.~Ma, E.~Donlon, S.~Dong, and A.~Rodriguez, ``Dense {{Tactile Force Estimation}} using {{GelSlim}} and inverse {{FEM}},'' in \emph{2019 {{International Conference}} on {{Robotics}} and {{Automation}} ({{ICRA}})}, May 2019, pp. 5418--5424.

\bibitem{hoganSeeingYourSkin2020}
F.~R. Hogan, M.~Jenkin, S.~{Rezaei-Shoshtari}, Y.~Girdhar, D.~Meger, and G.~Dudek, ``Seeing {{Through}} your {{Skin}}: {{Recognizing Objects}} with a {{Novel Visuotactile Sensor}},'' Dec. 2020.

\bibitem{hoganFingerSTSCombinedProximity2022}
F.~R. Hogan, J.-F. Tremblay, B.~H. Baghi, M.~Jenkin, K.~Siddiqi, and G.~Dudek, ``Finger-{{STS}}: {{Combined Proximity}} and {{Tactile Sensing}} for {{Robotic Manipulation}},'' \emph{IEEE Robotics and Automation Letters}, vol.~7, no.~4, pp. 10\,865--10\,872, Oct. 2022.

\bibitem{billardLearningHumans2016}
A.~G. Billard, S.~Calinon, and R.~Dillmann, ``Learning from {{Humans}},'' in \emph{Springer {{Handbook}} of {{Robotics}}}, B.~Siciliano and O.~Khatib, Eds.\hskip 1em plus 0.5em minus 0.4em\relax Cham: Springer International Publishing, 2016, pp. 1995--2014.

\bibitem{ablettSeeingAllAngles2021}
T.~Ablett, Y.~Zhai, and J.~Kelly, ``Seeing {{All}} the {{Angles}}: {{Learning Multiview Manipulation Policies}} for {{Contact-Rich Tasks}} from {{Demonstrations}},'' in \emph{Proceedings of the {{IEEE}}/{{RSJ International Conference}} on {{Intelligent Robots}} and {{Systems}} ({{IROS}}'21)}, Prague, Czech Republic, Sept. 2021.

\bibitem{liImmersiveDemonstrationsAre2023}
K.~Li, D.~Chappell, and N.~Rojas, ``Immersive {{Demonstrations}} are the {{Key}} to {{Imitation Learning}},'' Jan. 2023.

\bibitem{pervezNovelLearningDemonstration2017}
A.~Pervez, A.~Ali, J.-H. Ryu, and D.~Lee, ``Novel learning from demonstration approach for repetitive teleoperation tasks,'' in \emph{2017 {{IEEE World Haptics Conference}} ({{WHC}})}.\hskip 1em plus 0.5em minus 0.4em\relax Munich, Germany: IEEE, June 2017, pp. 60--65.

\bibitem{fischerComparisonTypesRobot2016}
K.~Fischer, \emph{et~al.}, ``A comparison of types of robot control for programming by {{Demonstration}},'' in \emph{2016 11th {{ACM}}/{{IEEE International Conference}} on {{Human-Robot Interaction}} ({{HRI}})}, Mar. 2016, pp. 213--220.

\bibitem{akgunRobotLearningDemonstration2011}
B.~Akgun and K.~Subramanian, ``Robot {{Learning}} from {{Demonstration}}: {{Kinesthetic Teaching}} vs. {{Teleoperation}},'' \emph{Technical Report}, 2011.

\bibitem{hoganImpedanceControlApproach1984}
N.~Hogan, ``Impedance {{Control}}: {{An Approach}} to {{Manipulation}},'' in \emph{1984 {{American Control Conference}}}, June 1984, pp. 304--313.

\bibitem{sicilianoForceControl2009}
B.~Siciliano, ``Force {{Control}},'' in \emph{Robotics: {{Modelling}}, {{Planning}} and {{Control}}}, ser. Advanced {{Textbooks}} in {{Control}} and {{Signal Processing}}, L.~Sciavicco, L.~Villani, and G.~Oriolo, Eds.\hskip 1em plus 0.5em minus 0.4em\relax London: Springer, 2009, pp. 363--405.

\bibitem{raibertHybridPositionForce1981}
M.~H. Raibert and J.~J. Craig, ``Hybrid {{Position}}/{{Force Control}} of {{Manipulators}},'' \emph{Journal of Dynamic Systems, Measurement, and Control}, vol. 103, no.~2, pp. 126--133, June 1981.

\bibitem{yamaguchiImplementingTactileBehaviors2017}
A.~Yamaguchi and C.~G. Atkeson, ``Implementing tactile behaviors using {{FingerVision}},'' in \emph{2017 {{IEEE-RAS}} 17th {{International Conference}} on {{Humanoid Robotics}} ({{Humanoids}})}, Nov. 2017, pp. 241--248.

\bibitem{yuanTactileMeasurementGelSight2014}
W.~Yuan, ``Tactile {{Measurement}} with a {{GelSight Sensor}},'' Master's thesis, Massachusetts Institute of Technology, 2014.

\bibitem{kimUVtacSwitchableUV2022}
W.~Kim, W.~D. Kim, J.-J. Kim, C.-H. Kim, and J.~Kim, ``{{UVtac}}: {{Switchable UV Marker-Based Tactile Sensing Finger}} for {{Effective Force Estimation}} and {{Object Localization}},'' \emph{IEEE Robotics and Automation Letters}, vol.~7, no.~3, pp. 6036--6043, July 2022.

\bibitem{jilaniTactileRecoveryShape2024}
A.~Jilani, ``Tactile recovery of shape from texture deformation,'' Master's thesis, McGill University, Jan. 2024.

\bibitem{bainFrameworkBehaviouralCloning1996}
M.~Bain and C.~Sammut, ``A {{Framework}} for {{Behavioural Cloning}},'' in \emph{Machine {{Intelligence}} 15}.\hskip 1em plus 0.5em minus 0.4em\relax Oxford University Press, 1996, pp. 103--129.

\bibitem{abbeelApprenticeshipLearningInverse2004}
P.~Abbeel and A.~Y. Ng, ``Apprenticeship learning via inverse reinforcement learning,'' in \emph{International {{Conference}} on {{Machine Learning}} ({{ICML}}'04)}.\hskip 1em plus 0.5em minus 0.4em\relax Banff, Alberta, Canada: ACM Press, 2004.

\bibitem{mandlekarWhatMattersLearning2021}
A.~Mandlekar, \emph{et~al.}, ``What {{Matters}} in {{Learning}} from {{Offline Human Demonstrations}} for {{Robot Manipulation}},'' in \emph{Conference on {{Robot Learning}}}, Nov. 2021.

\bibitem{zhangDeepImitationLearning2018}
T.~Zhang, \emph{et~al.}, ``Deep {{Imitation Learning}} for {{Complex Manipulation Tasks}} from {{Virtual Reality Teleoperation}},'' in \emph{Proceedings of the {{IEEE}} International Conference on Robotics and Automation ({{ICRA}}'18)}.\hskip 1em plus 0.5em minus 0.4em\relax Brisbane, QLD, Australia: IEEE, May 2018, pp. 5628--5635.

\bibitem{ablettLearningGuidedPlay2023}
T.~Ablett, B.~Chan, and J.~Kelly, ``Learning {{From Guided Play}}: {{Improving Exploration}} for {{Adversarial Imitation Learning With Simple Auxiliary Tasks}},'' \emph{IEEE Robotics and Automation Letters}, vol.~8, no.~3, pp. 1263--1270, Mar. 2023.

\bibitem{changImitationLearningObservation2023}
W.-D. Chang, S.~Fujimoto, D.~Meger, and G.~Dudek, ``Imitation {{Learning}} from {{Observation}} through {{Optimal Transport}},'' Oct. 2023.

\bibitem{orsiniWhatMattersAdversarial2021}
M.~Orsini, \emph{et~al.}, ``What {{Matters}} for {{Adversarial Imitation Learning}}?'' in \emph{Conference on {{Neural Information Processing Systems}}}, June 2021.

\bibitem{leeLearningForcebasedManipulation2015}
A.~X. Lee, H.~Lu, A.~Gupta, S.~Levine, and P.~Abbeel, ``Learning force-based manipulation of deformable objects from multiple demonstrations,'' in \emph{2015 {{IEEE International Conference}} on {{Robotics}} and {{Automation}} ({{ICRA}})}, May 2015, pp. 177--184.

\bibitem{abu-dakkaForcebasedVariableImpedance2018}
F.~J. {Abu-Dakka}, L.~Rozo, and D.~G. Caldwell, ``Force-based variable impedance learning for robotic manipulation,'' \emph{Robotics and Autonomous Systems}, vol. 109, pp. 156--167, Nov. 2018.

\bibitem{kormushevImitationLearningPositional2011}
P.~Kormushev, S.~Calinon, and D.~G. Caldwell, ``Imitation {{Learning}} of {{Positional}} and {{Force Skills Demonstrated}} via {{Kinesthetic Teaching}} and {{Haptic Input}},'' \emph{Advanced Robotics}, vol.~25, no.~5, pp. 581--603, Jan. 2011.

\bibitem{chebotarLearningRobotTactile2014}
Y.~Chebotar, O.~Kroemer, and J.~Peters, ``Learning robot tactile sensing for object manipulation,'' in \emph{2014 {{IEEE}}/{{RSJ International Conference}} on {{Intelligent Robots}} and {{Systems}}}, Sept. 2014, pp. 3368--3375.

\bibitem{limoyoLearningSequentialLatent2023}
O.~Limoyo, T.~Ablett, and J.~Kelly, ``Learning {{Sequential Latent Variable Models}} from~{{Multimodal Time Series Data}},'' in \emph{Intelligent {{Autonomous Systems}} 17}, ser. Lecture {{Notes}} in {{Networks}} and {{Systems}}, I.~Petrovic, E.~Menegatti, and I.~Markovi{\'c}, Eds.\hskip 1em plus 0.5em minus 0.4em\relax Cham: Springer Nature Switzerland, 2023, pp. 511--528.

\bibitem{hansenVisuotactileRLLearningMultimodal2022}
J.~Hansen, F.~Hogan, D.~Rivkin, D.~Meger, M.~Jenkin, and G.~Dudek, ``Visuotactile-{{RL}}: {{Learning Multimodal Manipulation Policies}} with {{Deep Reinforcement Learning}},'' in \emph{2022 {{International Conference}} on {{Robotics}} and {{Automation}} ({{ICRA}})}, May 2022, pp. 8298--8304.

\bibitem{huangRobotLearningDemonstration2020}
I.~Huang and R.~Bajcsy, ``Robot {{Learning}} from {{Demonstration}} with {{Tactile Signals}} for {{Geometry-Dependent Tasks}},'' in \emph{2020 {{IEEE}}/{{RSJ International Conference}} on {{Intelligent Robots}} and {{Systems}} ({{IROS}})}.\hskip 1em plus 0.5em minus 0.4em\relax Las Vegas, NV, USA: IEEE, Oct. 2020, pp. 8323--8328.

\bibitem{liSeeHearFeel2022}
H.~Li, \emph{et~al.}, ``See, {{Hear}}, and {{Feel}}: {{Smart Sensory Fusion}} for {{Robotic Manipulation}},'' in \emph{6th {{Annual Conference}} on {{Robot Learning}}}, Nov. 2022.

\bibitem{dasariRB2RoboticManipulation2021}
S.~Dasari, \emph{et~al.}, ``{{RB2}}: {{Robotic Manipulation Benchmarking}} with a {{Twist}},'' in \emph{Thirty-Fifth {{Conference}} on {{Neural Information Processing Systems Datasets}} and {{Benchmarks Track}} ({{Round}} 2)}, Oct. 2021.

\bibitem{figueroaEasykinestheticrecording2023}
N.~Figueroa, ``Easy-kinesthetic-recording,'' https://github.com/nbfigueroa/easy-kinesthetic-recording, Oct. 2023.

\bibitem{villaniForceControl2016}
L.~Villani and J.~De~Schutter, ``Force {{Control}},'' in \emph{Springer {{Handbook}} of {{Robotics}}}, ser. Springer {{Handbooks}}, B.~Siciliano and O.~Khatib, Eds.\hskip 1em plus 0.5em minus 0.4em\relax Cham: Springer International Publishing, 2016, pp. 195--220.

\bibitem{opencv_library}
G.~Bradski, ``The {{OpenCV}} library,'' \emph{Dr. Dobb's Journal of Software Tools}, 2000.

\bibitem{tsaiNewTechniqueFully1989}
R.~Tsai and R.~Lenz, ``A new technique for fully autonomous and efficient {{3D}} robotics hand/eye calibration,'' \emph{IEEE Transactions on Robotics and Automation}, vol.~5, no.~3, pp. 345--358, June 1989.

\bibitem{limoyoSelfCalibrationMobileManipulator2018}
O.~Limoyo, T.~Ablett, F.~Mari{\'c}, L.~Volpatti, and J.~Kelly, ``Self-{{Calibration}} of {{Mobile Manipulator Kinematic}} and {{Sensor Extrinsic Parameters Through Contact-Based Interaction}},'' in \emph{2018 {{IEEE International Conference}} on {{Robotics}} and {{Automation}} ({{ICRA}})}, May 2018, pp. 1--8.

\bibitem{Polymetis2021}
Y.~Lin, A.~S. Wang, G.~Sutanto, A.~Rai, and F.~Meier, ``Polymetis,'' https://facebookresearch.github.io/fairo/polymetis/, 2021.

\bibitem{heDeepResidualLearning2016}
K.~He, X.~Zhang, S.~Ren, and J.~Sun, ``Deep {{Residual Learning}} for {{Image Recognition}},'' in \emph{Proceedings of the {{IEEE}} Conference on Computer Vision and Pattern Recognition ({{CVPR}}'16)}.\hskip 1em plus 0.5em minus 0.4em\relax Las Vegas, NV, USA: IEEE, June 2016, pp. 770--778.

\bibitem{levineEndtoendTrainingDeep2016}
S.~Levine, C.~Finn, T.~Darrell, and P.~Abbeel, ``End-to-end training of deep visuomotor policies,'' \emph{Journal of Machine Learning Research}, vol.~17, no.~39, pp. 1--40, 2016.

\bibitem{dehaanCausalConfusionImitation2019}
P.~{de Haan}, D.~Jayaraman, and S.~Levine, ``Causal {{Confusion}} in {{Imitation Learning}},'' in \emph{Advances in {{Neural Information Processing Systems}} ({{Neurips}}'19)}, 2019, pp. 11\,693--11\,704.

\bibitem{limoyoHeteroscedasticUncertaintyRobust2020}
O.~Limoyo, B.~Chan, F.~Mari{\'c}, B.~Wagstaff, A.~R. Mahmood, and J.~Kelly, ``Heteroscedastic {{Uncertainty}} for {{Robust Generative Latent Dynamics}},'' \emph{IEEE Robotics and Automation Letters}, vol.~5, no.~4, pp. 6654--6661, Oct. 2020.

\end{thebibliography}

\begin{IEEEbiography}[{\includegraphics[width=1in,height=1.25in,clip,keepaspectratio]{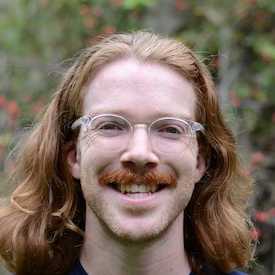}}]{Trevor Ablett} received his Bachelor of Engineering in Mechatronics along with a Bachelor of Arts in Psychology from McMaster University, Hamilton, Canada in 2015. 
He is currently a Ph.D.\ Candidate in the Space and Terrestrial Autonomous Robotics (STARS) Laboratory at the University of Toronto Robotics Institute, Toronto, Canada.
His research interests include imitation and reinforcement learning, robotic manipulation, and improving the robustness and efficiency of learning-based robotics.
\end{IEEEbiography}
\begin{IEEEbiography}[{\includegraphics[width=1in,height=1.25in,clip,keepaspectratio]{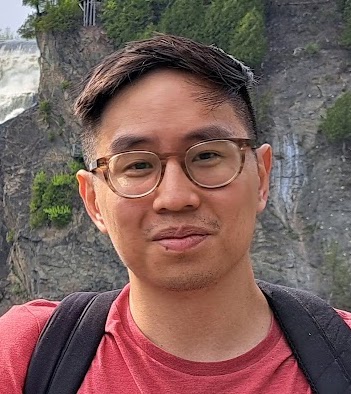}}]
{Oliver Limoyo} received his Ph.D. degree from the University of Toronto in 2024. 
He is currently a research scientist at Waabi. 
His research interests include the application of generative and self-supervised learning to robotics problems, and multimodal sensing for manipulation.
\end{IEEEbiography}
\begin{IEEEbiography}[{\includegraphics[width=1in,height=1.25in,clip,keepaspectratio]{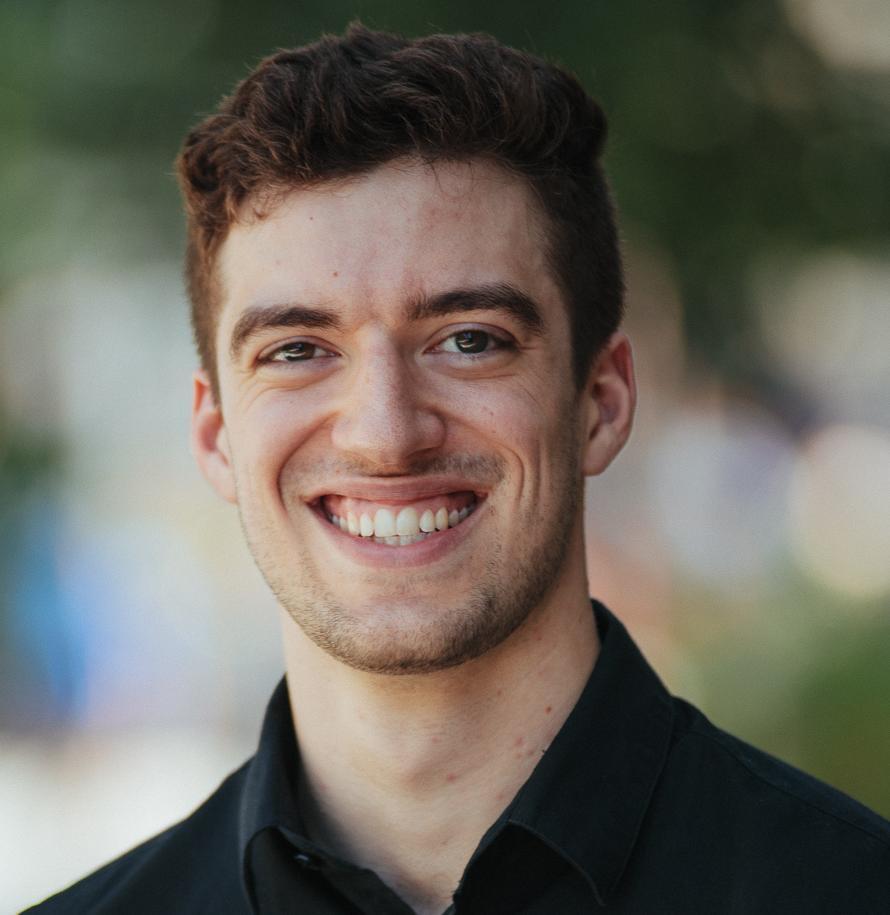}}]
{Adam Sigal} received his B.S.\ from Université de Montréal in Computer Science in 2019, and his M.Eng.\ in Electrical Engineering from McGill University in 2022. After graduating, he joined the Samsung AI Center in Montr\'eal as a machine learning research engineer. His research has focused on autonomous robot navigation, human-robot interaction, robot manipulation with visuotactile feedback, and more recently, embodied applications of generative models. 
\end{IEEEbiography}
\begin{IEEEbiography}[{\includegraphics[width=1in,height=1.25in,clip,keepaspectratio]{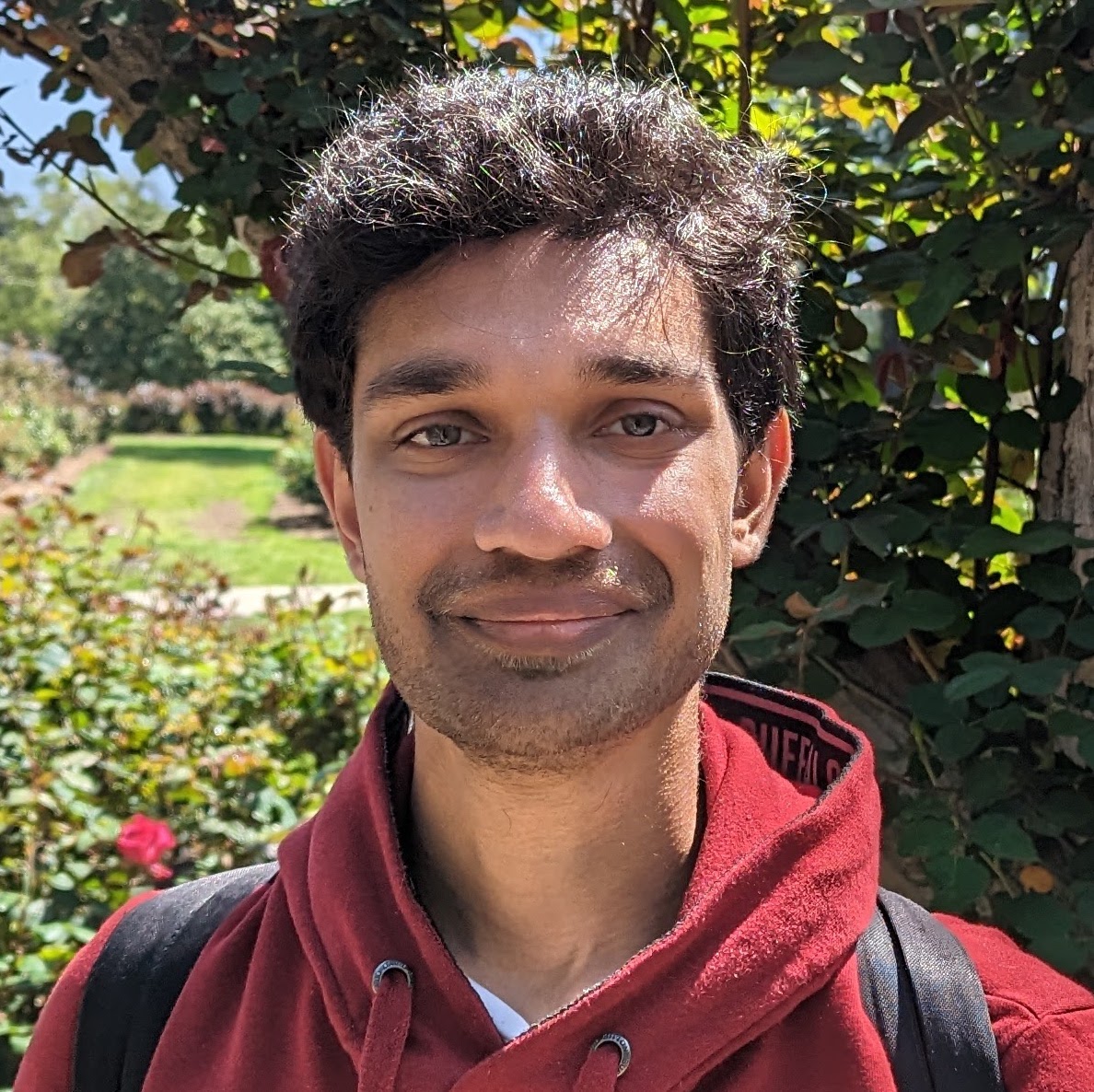}}]
{Affan Jilani} received his Bachelor of Science degree in Computer Science and his M.Sc. degree from McGill University in 2020 and 2024, respectively.
His research interests are focused on solving computer vision problems, and include shape recovery and applications of computer vision for robotics. 
\end{IEEEbiography}
\begin{IEEEbiography}[{\includegraphics[width=1in,height=1.25in,clip,keepaspectratio]{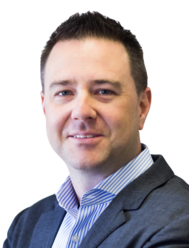}}]{Jonathan Kelly} received his Ph.D.\ degree in Computer Science from the University of Southern California, Los Angeles, USA, in 2011. 
From 2011 to 2013 he was a postdoctoral associate in the Computer Science and Artificial Intelligence Laboratory at the Massachusetts Institute of Technology, Cambridge, USA. 
He is currently an associate professor and director of the Space and Terrestrial Autonomous Robotic Systems (STARS) Laboratory at the University of Toronto Institute for Aerospace Studies, Toronto, Canada. 
Prof.\ Kelly holds the Tier II Canada Research Chair in Collaborative Robotics. 
His research interests include perception, planning, and learning for interactive robotic systems.
\end{IEEEbiography}
\begin{IEEEbiography}[{\includegraphics[width=1in,height=1.25in,clip,keepaspectratio]{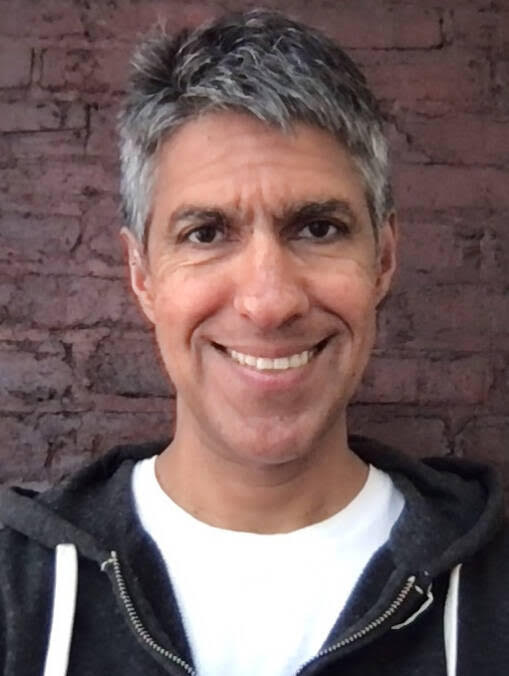}}]
{Kaleem Siddiqi} received his BS degree from Lafayette College in 1988 and his MS and PhD degrees from Brown University in 1990 and 1995, respectively, all in the field of electrical engineering. He is currently a Professor at the School of Computer Science at McGill University where he holds an FRQS Dual Chair in Health and Artificial Intelligence. 
He is also a member of McGill’s Centre for Intelligent Machines, an associate member of McGill’s Department of Mathematics and Statistics, MILA - the Qu\'ebec AI Institute, and the Goodman Cancer Centre. 
He presently serves as Field Chief Editor of the journal Frontiers in Computer Science. 
Before moving to McGill in 1998, he was a postdoctoral associate in the Department of Computer Science at Yale University (1996-1998) and held a position in the Department of Electrical Engineering at McGill University (1995-1996). 
More recently he was also a visiting professor and consultant at the Samsung AI Centre in Montr\'eal (2021-2023). 
His research interests are in computer vision, robotics, image analysis, biological shape, neuroscience and medical imaging.
He is a member of Phi Beta Kappa, Tau Beta Pi, and Eta Kappa Nu. He is a senior member of the IEEE.
\end{IEEEbiography}
\begin{IEEEbiography}[{\includegraphics[width=1in,height=1.25in,clip,keepaspectratio]{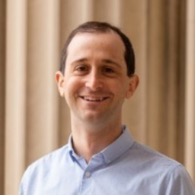}}]
{Francois Hogan} 

Francois Hogan received his B.S. and M.S.\ degrees in Mechanical Engineering from McGill University in 2013 and 2015 respectively.  He received his Ph.D.\ in Mechanical Engineering from the Massachusetts Institute of Technology in 2019.  Francois is currently Research Scientist at Meta. His primary research focuses on exploiting tactile and visual feedback to enable robots to reliably manipulate their environment. Francois was a member of team MIT-Princeton's 1st place finish 2017 at the Amazon Robotics Challenge's Stowing task, is the recipient of the MIT Presidential Fellowship, was awarded the 2018 Amazon Robotics Best Systems Paper, and was a 2021 Finalist for Best Manipulation paper at ICRA 2020.
\end{IEEEbiography}
\begin{IEEEbiography}[{\includegraphics[width=1in,height=1.25in,clip,keepaspectratio]{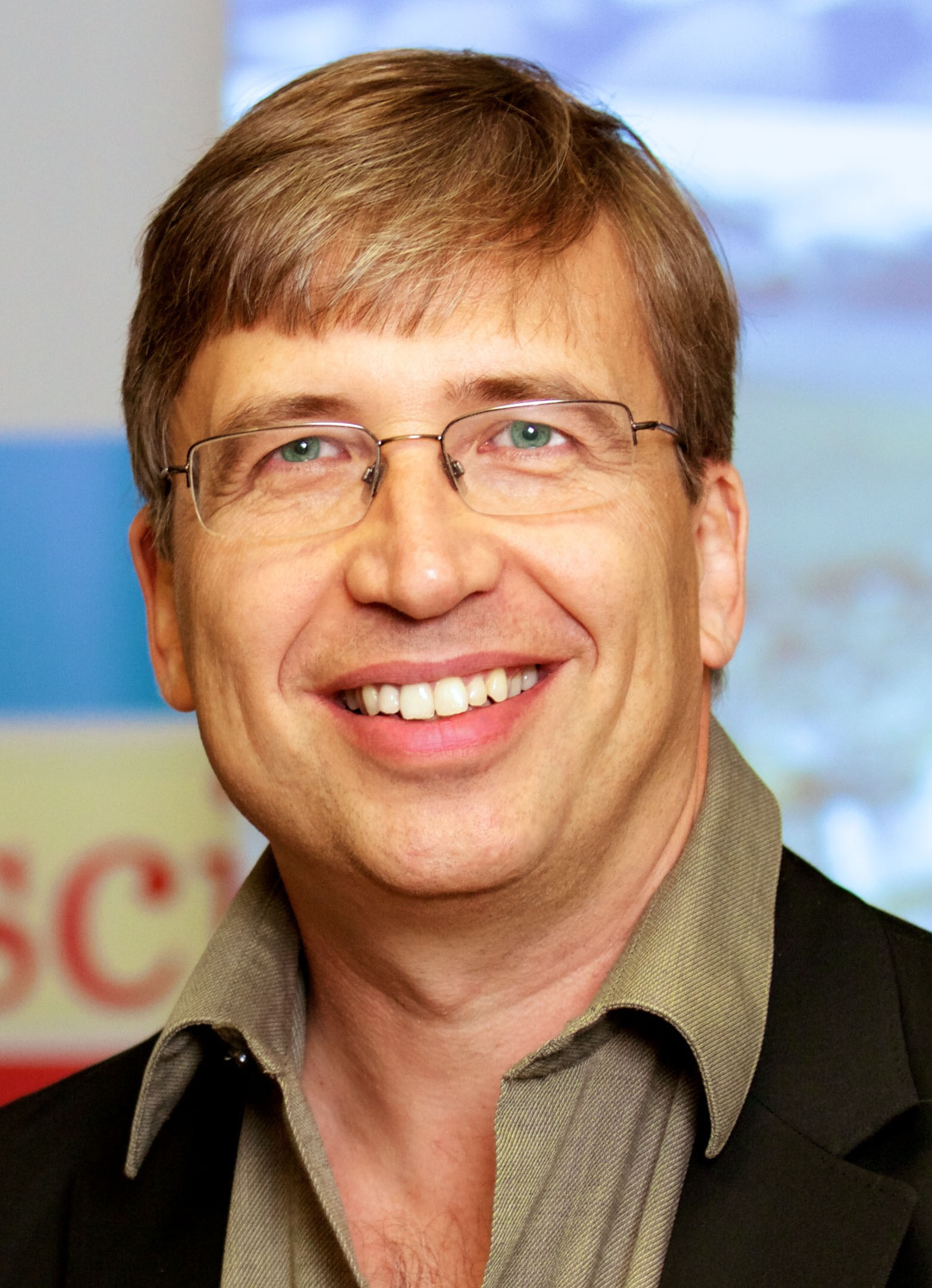}}]
{Gregory Dudek} is a professor and former director of the School of Computer Science at McGill Research Centre for Intelligent Machines (CIM), Research Director of Mobile Robotics Lab at McGill, and an Associate Member of the Dept. of Electrical Engineering. In 2008 he was made James McGill Chair. Since 2012 he has been the Scientific Director of the NSERC Canadian Field Robotics Network (NCFRN). He was the Lab Head and VP, Research of Samsung’s AI Center in Montreal from 2019-2023. 
He has chaired and been otherwise involved in numerous national and international conferences and professional activities concerned with Robotics, Machine Sensing and Computer Vision. His research interests include perception for mobile robotics, navigation and position estimation, environment and shape modelling, computational vision and collaborative filtering.
\end{IEEEbiography}

\end{document}